\documentclass[sigconf]{acmart}
\AtBeginDocument{%
  \providecommand\BibTeX{{%
    \normalfont B\kern-0.5em{\scshape i\kern-0.25em b}\kern-0.8em\TeX}}}

\setcopyright{acmcopyright}
\copyrightyear{2018}
\acmYear{2018}
\acmDOI{10.1145/1122445.1122456}

\acmConference[Woodstock '18]{Woodstock '18: ACM Symposium on Neural
  Gaze Detection}{June 03--05, 2018}{Woodstock, NY}
\acmBooktitle{Woodstock '18: ACM Symposium on Neural Gaze Detection,
  June 03--05, 2018, Woodstock, NY}
\acmPrice{15.00}
\acmISBN{978-1-4503-XXXX-X/18/06}

\usepackage{gensymb}
\usepackage{multirow}
\usepackage{makecell}
\usepackage{subfigure}
\usepackage{graphicx}
\usepackage{multirow}

\begin{document}

\title{CubeLearn: End-to-end Learning for Human Motion Recognition from Raw mmWave Radar Signals}



\author{Peijun Zhao}
\affiliation{%
  \institution{University of Oxford}
  \city{Oxford}
  \country{United Kingdom}
}
\author{Chris Xiaoxuan Lu}
\affiliation{%
  \institution{University of Edinburgh}
  \city{Edinburgh}
  \country{United Kingdom}}

\author{Bing Wang}
\affiliation{%
  \institution{University of Oxford}
  \city{Oxford}
  \country{United Kingdom}
}
\author{Niki Trigoni}
\affiliation{%
  \institution{University of Oxford}
  \city{Oxford}
  \country{United Kingdom}
}
\author{Andrew Markham}
\affiliation{%
  \institution{University of Oxford}
  \city{Oxford}
  \country{United Kingdom}
}

\renewcommand{\shortauthors}{Zhao et al.}

\begin{abstract}
  mmWave FMCW radar has attracted huge amount of research interest for human-centered applications in recent years, such as human gesture/activity recognition. Most existing pipelines are built upon conventional Discrete Fourier Transform (DFT) pre-processing and deep neural network classifier hybrid methods, with a majority of previous works focusing on designing the downstream classifier to improve overall accuracy. In this work, we take a step back and look at the pre-processing module. To avoid the drawbacks of conventional DFT pre-processing, we propose a learnable pre-processing module, named CubeLearn, to directly extract features from raw radar signal and build an end-to-end deep neural network for mmWave FMCW radar motion recognition applications. Extensive experiments show that our CubeLearn module consistently improves the classification accuracies of different pipelines, especially benefiting those previously weaker models. We provide ablation studies on initialization methods and structure of the proposed module, as well as an evaluation of the running time on PC and edge devices. This work also serves as a comparison of different approaches towards data cube slicing. Through our task agnostic design, we propose a first step towards a generic end-to-end solution for radar recognition problems.
\end{abstract}




\maketitle
\section{Introduction}

\begin{figure}[t]
    \centering
    \includegraphics[width=0.45\textwidth]{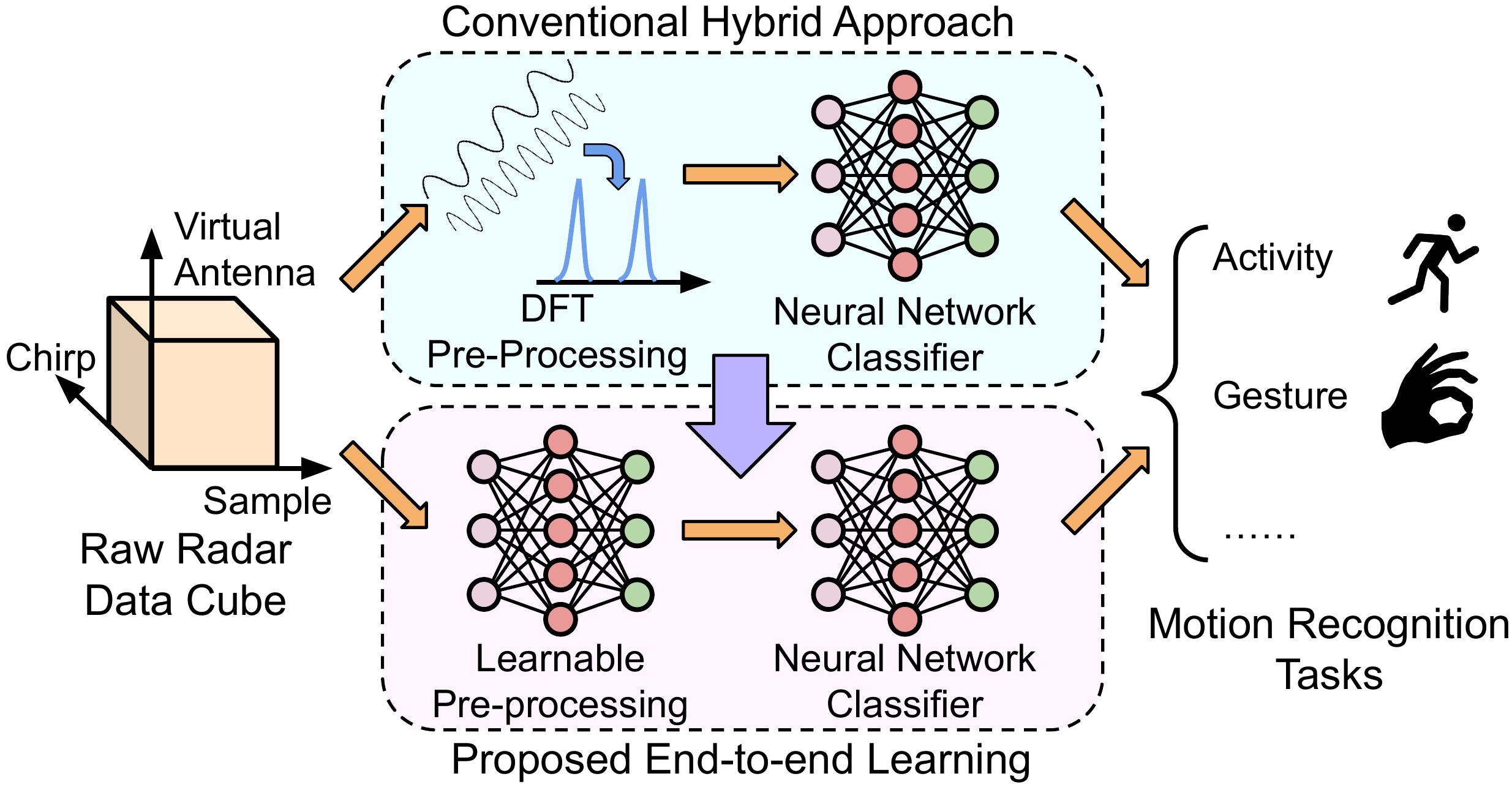}
    \caption{We propose CubeLearn, a learnable pre-processing module to replace conventional DFT and build end-to-end deep neural networks for mmWave FMCW radar motion recognition tasks.}
    \label{fig:opening}
\end{figure}

Millimeter wave (mmWave) frequency-modulated continuous-wave (FMCW) radar was mainly used on high-end cars and military vehicles many years ago, due to its bulky size and high cost. With the recent development in low cost single-chip mmWave radar, (e.g., TI mmWave sensors \cite{TImmwave} and Infineon mmWave sensors including Google Soli \cite{lien2016soli}), more and more possibilities have been explored in indoor applications, such as vital sign monitoring \cite{singh2020multi}, gesture recognition \cite{palipana2021pantomime, wang2016interacting}, fall detection \cite{jin2020mmfall, sun2019privacy, wang2020millimetre}, and human activity recognition (HAR) \cite{singh2019radhar, ahuja2021vid2doppler}. As mmWave radars are less obtrusive than cameras, they have significant potential for domestic applications, where concerns about privacy dominate.

Typical methods for FMCW radar gesture or activity recognition follow a two-stage pipeline: radar signal pre-processing, typically with Discrete Fourier Transform (DFT), and a data-driven classifier for task-specific recognition purposes. In the classical radar processing chain, there are a number of different levels of data representations, ranging from Range Profile signatures, heatmaps, to point clouds. 

For gesture recognition and activity recognition tasks, heatmaps and point clouds are the most widely adopted data representations \cite{ahmed2021hand,li2019survey}. Point cloud generation relies on hand crafted parameters, and generating point clouds requires receiver array which is not available in some radar configurations (e.g., SISO radar). Besides, point cloud can hardly capture find-grained movements like finger gestures, which makes it a less universal approach. As a result, most previous works are based on heatmap + neural network classifier hybrid pipelines.

DFT serves as a very efficient way to extract Range, Doppler and Angle-of-Arrival (AoA) information from the raw radar signal to generate different types of heatmaps. However, it also has limitations including: (1) it uses preset, equally spaced transformation bases, which could be sub-optimal; (2) limited resolution; (3) heatmaps contain a large portion of redundant information. To overcome such limitations, researchers have been trying to use end-to-end deep neural networks for radar applications in recent years. The difficulty here lies in that the raw radar ADC samples are complex values corresponding to the downmixed baseband, and both the magnitude and the phase are important. In previous works on end-to-end learning for radar applications, either only the real part is used \cite{stadelmayer2021data, ye2019using}, which loses some information; or the real part and the imaginary part are treated as two channels \cite{zhao2020end}, which loses the physical meaning, and using a specific network structure could make it less universal.

In this work, we explore replacing DFT pre-processing with a learnable pre-processing module named CubeLearn to improve motion classification accuracy, which is evaluated on human gesture and activity recognition tasks. To the authors' best knowledge, this is the first work to use stacked complex linear layers to replace DFT pre-processing for FMCW radar pipelines, and the first in-depth evaluation of a purely end-to-end radar network. Our intuition is that by making the entire network learnable and data-driven, we can allow the network to better focus on important radar features, leading to higher overall accuracy. In particular, significant performance improvements are observed for those previously weaker models, indicating that our proposed CubeLearn module could be especially helpful for low-end radars (e.g., with single Tx/Rx) to achieve recognition performance close to a more sophisticated radar. Besides, compared to previous end-to-end works where the proposed methods are designed for specific applications, our work proposes a more generic method for different radar recognition tasks.

Our contributions in this work are listed as follows:
\begin{itemize}
    \item We propose CubeLearn, a learnable pre-processing module to replace the conventional DFT pre-processing, allowing the network to focus on task-specific radar signatures.
    \item Through extensive evaluation, we show that our proposed method can increase the accuracy of different gesture/activity recognition pipelines and accelerate the training process.
    \item We also provide a detailed, in-depth exploration of the impact of different pre-processing and classifier combinations, in terms of task accuracy and computational load on both PC and Raspberry Pi devices.
    \item We release the code of the proposed CubeLearn module and our own collected dataset to encourage further research on this topic\footnote{\url{https://github.com/zhaoymn/cubelearn}}.
\end{itemize}

In the rest of the paper, we first discuss related work, mmWave radar background and commonly seen pipelines. Then we present our proposed CubeLearn module in Section~\ref{sec:learnable}. In Section~\ref{sec:experiment_setup} we introduce the dataset collection and experiment configuration, followed by evaluation in Section~\ref{sec:evaluation}. In the remaining sections we provide extensive ablation study and complexity analysis, and Section~\ref{sec:conclusion} finally concludes the paper.

\section{\label{sec:related_works}Related Work}

\subsection{mmWave Radar Heatmap Based Human Gesture/Activity Recognition}

\begin{table}[h]
    \centering
    \scriptsize
    \vspace{-10pt}
    \begin{tabular}{c|c|c|c}
        \hline
        \multicolumn{2}{c|}{\textbf{Pre-processing}} & \textbf{Classifier} & \textbf{Related Works}\\
        \hline
        1D Profile & Range, Angle, Doppler Profile & LSTM & \cite{zheng2019hand}\\
        \hline
        \multirow{7}{*}{2D Heatmap} & Doppler-Time  & 2DCNN & \cite{dekker2017gesture, wu2018dynamic, ahuja2021vid2doppler, jiang2021recognition, zhang2018real}\\
         & Range-Doppler  & 2DCNN-LSTM & \cite{wang2016interacting, wang2019ts}\\
         & Range-Doppler  & 2DCNN-TCN & \cite{9381994}\\
         & Range-Doppler  & ANN & \cite{goswami2019real}\\
         & Range-Doppler + Range-Angle  & 2DCNN-LSTM & \cite{yu2020mmwave}\\
         & Range-Time + Doppler-Time  & 2DCNN & \cite{huang2020rd}\\
         & Range-Time + Angle-Time  & 2DCNN & \cite{wang2019two}\\
        \hline
        \multirow{2}{*}{3D Heatmap} & Range-Doppler-Time  & 3DCNN & \cite{chen2019gesture, hazra2019short}\\
         & Range-Doppler-Angle & 2DCNN + LSTM & \cite{hazra2019radar}\\
        \hline
    \end{tabular}
    \caption{Heatmap based mmWave FMCW radar gesture and activity recognition.}
    \label{tab:gesture_activity_recognition}
    \vspace{-20pt}
\end{table}

mmWave radar based human gesture/activity recognition is one of the most studied areas in mmWave sensing.
Main-stream recognition methods are based on 1D/2D/3D heatmaps (sometimes 1D representations are called `profile'). Recent works on mmWave radar heatmap based human gesture/activity recognition are summarized in TABLE~\ref{tab:gesture_activity_recognition}. As introduced in Section~\ref{sec:background}, conventional DFT pre-processing is used to extract Range, Doppler and AoA information from the raw radar data cube. In most cases the heatmap data is directly used as neural network input \cite{dekker2017gesture, wu2018dynamic, ahuja2021vid2doppler, jiang2021recognition, zhang2018real, wang2019ts, 9381994, yu2020mmwave, huang2020rd, wang2019two, chen2019gesture, hazra2019short}, as they are fixed-sized and can be processed similarly to images. Other works manually extract features and feed them into the downstream classifier \cite{zheng2019hand, goswami2019real, hazra2019radar}.

There are many types of 2D and 3D heatmaps, such as the Range-Doppler heatmap used in \cite{wang2016interacting, wang2019ts,9381994,goswami2019real,yu2020mmwave}. The frame dimension can be used as a separate dimension for the neural network to extract temporal information\cite{wang2016interacting, wang2019ts, 9381994, yu2020mmwave,hazra2019radar}, or use together with other dimensions to be processed as part of the heatmap \cite{dekker2017gesture, wu2018dynamic, franceschini2020hand,chen2019gesture, hazra2019short}. In our work we consider different heatmap based pipelines in previous works, together with other possible combinations, as our baselines.

\subsection{End-to-end mmWave Radar Gesture and Activity Recognition Methods}
Researchers have also been trying to build end-to-end neural networks for radar applications. The main difficulty of designing an end-to-end structure is to handle the complex input. The magnitude and phase information are both important in the raw IF signal, as the magnitude contains information regarding target distance, and the phase information can be used for velocity estimation and AoA estimation. 

Sakamoto et al. proposed converting the I/Q data received by CW radar receiver into an image for hand gesture recognition with convolutional neural networks \cite{sakamoto2018hand}. Zhao et al. proposed treating the real and imaginary part as two separate channels which is processed similar to images with real-valued convolutional layers for human activity classification \cite{zhao2020end}. 

There are also some works that use real-valued input. Stadelmayer and Santra proposed using a parametric convolutional neural network (2D Sinc Filter and 2D Wavelet Filter) \cite{stadelmayer2021data} to extract Range and Doppler information from raw radar data. Ye et al. proposed using two real-valued convolutional layers with Fourier initialization for human-activity classification \cite{ye2019using, ye2020human} on Continuous-wave radar data. However, their methods lose half of the information from the imaginary part. 

To the authors best knowledge, this work is the first to use stacked complex linear layers to replace the conventional DFT pre-processing to directly extract information from raw mmWave FMCW radar data, and to build end-to-end deep neural network together with downstream deep classifier which directly takes raw radar complex data cube as input for recognition tasks.
\section{\label{sec:background}mmWave FMCW Radar Background}

In a typical FMCW radar configuration, a frame consists of multiple chirps, which are short periods of signals whose frequency increases linearly with time. During each chirp, a number of analog-digital converter (ADC) samples are taken at each receiver. We use Texas Instruments IWR6843 mmWave radar in this work, which features a 3 transmitter, 4 receiver MIMO antenna array, but we note that our architecture is easily generalizable to different antenna configurations. For each chirp, typically a single transmitter is activated (with the TDM-MIMO \cite{rambach2017mimo} principle), and the received signals at each receiver antenna are mixed with the transmitting signal followed by low-pass filters to produce an intermediate frequency (IF) signal that is further sampled by ADC. For each frame, mmWave radar generates a raw data cube, with the three axes namely corresponding to: samples in a chirp, chirp loops, and Tx-Rx pairs. 

In conventional processing, DFT is used to extract range, Doppler and AoA information along the 3 dimensions of the raw radar data cube, respectively. 
As the transmitting frequency increases linearly with time during a chirp, the reflection from a target at a certain distance introduces a corresponding frequency component in the IF signal, which equals to the round trip time multiplied by the frequency slope of the chirp. We can estimate the distance to different targets by extracting different frequency components with DFT, which produces a `Range Profile'. The phase of a peak corresponding to a certain target in the Range profile changes across chirps if the target is moving, due to the slight round-trip distance change, and we can apply another DFT across chirps to extract the phase variation in order to infer the radial velocity of the object, which is often called `Doppler DFT'. Furthermore, the AoA of the target introduces phase differences at the antennas in the receiver array because of the slight difference in the total length the signal travels. This can also be estimated with DFT, which is called `Angle DFT'. Transforming the information from the sampled time/space domain to the frequency domain is the key for extracting range, Doppler and AoA information in FMCW radar data processing.

\section{Gesture and activity recognition models\label{sec:pipelines}}
As discussed in the previous section, the mmWave radar is able to produce a raw data cube for each frame, with 3 axes as sample, chirp and receiver array, respectively, and range, Doppler and AoA information can be extracted from these three dimensions. To represent the information on these three dimensions, we often adopt heatmaps as a straightforward data representation. Peaks in the heatmaps correspond to detected targets. For example, a peak in the 2D Range-Doppler heatmap represents a target at a certain distance with a certain radial velocity. Since in this work we are studying gesture/activity recognition, which typically lasts several frames, we have an additional frame (time) dimension. As a result, with different combinations of information from three dimensions in the data cube, together with the frame dimension, we supply 2D, 3D or 4D data as input to the downstream neural network classifier.

\begin{figure}[ht]\centering
\subfigure[CNN classifier]{
\begin{minipage}{0.45\textwidth}\centering
\vspace{-5pt}
\includegraphics[width=\textwidth]{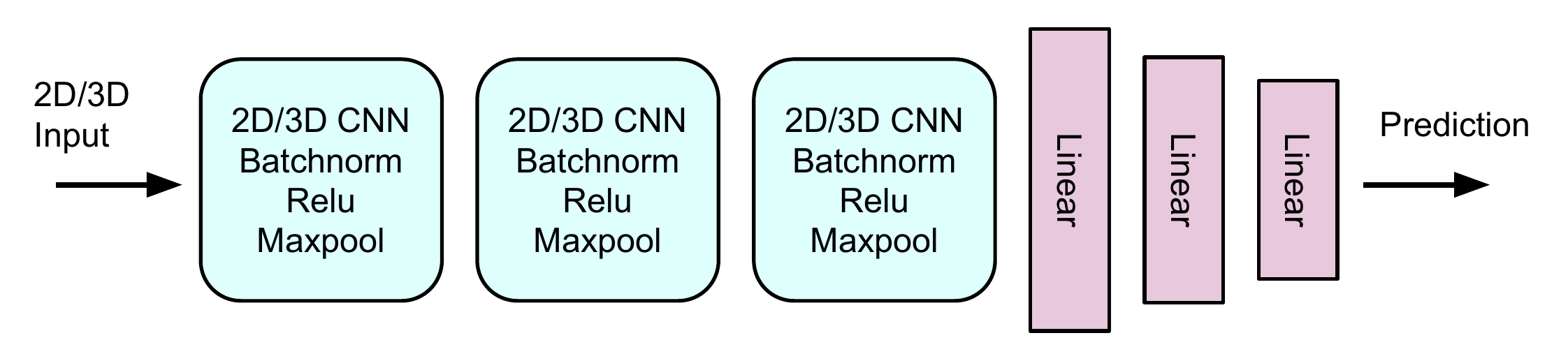}
\end{minipage}
}
\subfigure[CNN-LSTM classifier]{
\begin{minipage}{0.45\textwidth}\centering
\vspace{-15pt}
\includegraphics[width=\textwidth]{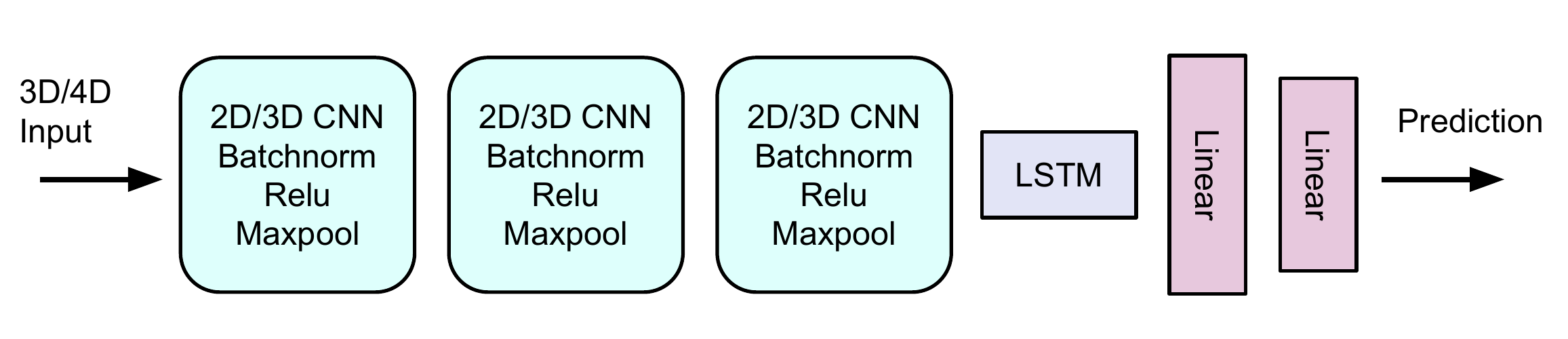}
\end{minipage}
}
\vspace{-15pt}
\caption{Two types of downstream classifier architectures we use in this work.}
\vspace{-5pt}
\label{fig:classifier}
\end{figure}

In this study, we evaluate two types of commonly adopted neural network classifiers: CNN and CNN-LSTM. The architecture of the two types of classifiers is shown in Fig.~\ref{fig:classifier}. We use three convolutional layers, each followed by batch normalization, activation (ReLU), and max-pooling. The output of the convolutional layers is flattened, followed by either LSTM and two fully connected layers, or three fully connected layers, to produce the prediction.

\begin{table}[h]
    \centering
    \scriptsize
    \begin{tabular}{c|c|c}
        \hline
        Data Cube Slicing & Pre-processing & Classifier\\
        \hline
        first Tx-Rx pair and first chirp & Range-Time & 2DCNN \\
        first Tx-Rx pair, range aggregation & Doppler-Time & 2DCNN \\
        first chirp, range aggregation & Angle-Time & 2DCNN \\
        first Tx-Rx pair & Range-Doppler-Time & 3DCNN/2DCNN-LSTM \\
        first chirp & Range-Angle-Time & 3DCNN/2DCNN-LSTM \\
        whole cube, range aggregation & Doppler-Angle-Time & 3DCNN/2DCNN-LSTM \\
        whole cube & Range-Doppler-Angle-Time & 3DCNN-LSTM \\
        \hline
    \end{tabular}
    \caption{Heatmap based mmWave FMCW radar gesture and activity recognition.}
    \label{tab:heatmap_classifier}
    \vspace{-15pt}
\end{table}

The frame can sometimes be used as one dimension in the heatmap, e.g., in a Range-Doppler-Time Cube, which can be fed into 3DCNN classifier, or used as sequence timestamps, to act as input into LSTM. We use TABLE~\ref{tab:heatmap_classifier} to summarize possible data slicing, pre-processing and downstream classifier combinations. Note that since we are using FMCW radar, for Doppler-Time, Angle-Time and Doppler-Angle-Time pre-processing, we still extract Range information first, and then estimate Doppler and/or AoA information for each Range bin, then sum along the Range axis. Some of the combinations have appeared in previous works, as we discuss in Section~\ref{sec:related_works}, and in these works the selection of the pipeline is largely based on the target application and radar configuration.

\section{\label{sec:learnable}Learnable pre-processing}
\subsection{Problems with Conventional DFT pre-processing}
For raw radar signals, the information of the targets is not encoded in certain data points, but rather hidden in the phase and magnitude variation of the signal. For example, the distances to the targets are encoded in the frequency components of a single chirp, and the Doppler and AoA information are hidden in the phase variation, either from time domain (chirps) or spatial domain (receiver array).

To expose such information, the pre-processing is often performed with conventional signal processing techniques, typically through DFT, as discussed before. 
The DFT can be viewed as a linear transformation of the raw radar data from the time/spatial domain to the frequency domain with equally spaced Fourier bases. By transforming to the frequency domain, the information corresponding to the targets is directly exposed. The pre-processed output can be largely regarded as 2D/3D images to be fed into downstream neural networks based on architectures widely explored in the computer vision domain. Note that though DFT is a reversible operation, in previous pipelines the phase information of the transformed data is lost before feeding into the downstream classifier because of the modulus operation.

Though conventional DFT pre-processing is simple, efficient and straightforward, there are three issues which impact its utility:

\noindent \textbf{Lack of adaptive transformation bases.} All the Fourier transformation bases are equally spaced and fixed, and the transformed data points might not best represent the target. For example, if we have a radar configuration where the distance between two data points in the Range Profile is $10cm$. The position of the actual target could be $35cm$ which falls between two sample points ($30cm$ and $40cm$) in the Range Profile. Although the adjacent data points may contain the target information, the representation is sub-optimal and the classifier will need to interpolate adjacent data points.

\noindent \textbf{Limited resolution.} Another problem is that DFT has inherent resolution limitation, which makes it hard for the DFT transformation output to represent the target properties on certain tasks. For example, the Range resolution of mmWave radar is typically $\textgreater 5cm$, which makes it difficult to distinguish two fingers in hand gesture recognition. Also, the resolution limitations of Doppler and AoA make it hard to capture fine-grained movements, e.g., movements of two adjacent fingers. Note that such problem could not be simply fixed by `Zero Padding' in time domain, as `Zero Padding' in time domain corresponds to interpolation in frequency domain, which does not increase the resolution. 

\noindent \textbf{Decoupled with task-oriented information.} Last but not least, as conventional DFT is a universal solution for extracting frequency domain information, and is not designed specifically for certain tasks, the part of the conventional DFT output that we are actually interested in only occupies a fraction of the whole pre-processing output, depending on specific tasks. 
For example in the Range dimension, in tasks like hand-gesture recognition, the movement is typically within 5cm, and only a couple of points in the Range Profile contain the information of the gesture performed (excluding multipath).
Even for activity recognition tasks where the user may perform actions across the longer range (e.g., 50cm), and such action can be captured with tens of data points in Range Profile, it's still only a small portion of the whole Range spectrum. The Doppler and AoA dimensions DFT also have similar limitations. The non-informative part of the output may contain multipath reflections, noise and environmental reflections. Sending this information into the downstream neural network may lead to much higher computational cost, and even worse, make the neural network learn fake features and overfit to the training data. 

It is possible to have a specific DFT for each task, with different FFT lengths, non-uniform frequency spacings or frequency masks such that the domain transformation can focus more on relevant features corresponding to specific tasks. However, this means that we must choose specific parameters for each task, which relies on the experience of the developer and is very time-consuming to tune these hyper-parameters.

\subsection{CubeLearn Module Architecture}

To deal with the above-mentioned limitations of conventional DFT pre-processing, in this work we present CubeLearn module to replace the DFT pre-processing, which can be trained together with the downstream classifier in an end-to-end way, and aim to learn the best domain transformation for a specific task in a data-driven manner. 

A neural network does not suffer from the limitations of conventional DFT. Through end-to-end training, the network is able to learn the transformation bases that best suit the target task. In addition, a neural network does not have the resolution limitation as in conventional DFT signal processing. In fact, recent research has shown that super-resolution of Range, Doppler and AoA can be achieved through deep neural networks for raw radar signals~\cite{schutz2020neural, gall2020spectrum}. Furthermore, the pre-processing module could be trained to extract more informative features, rather than noise and environmental reflections, potentially carrying more useful information to the downstream classifier.

\begin{figure}[h]
    \vspace{-5pt}
    \centering
    \includegraphics[width=0.45\textwidth]{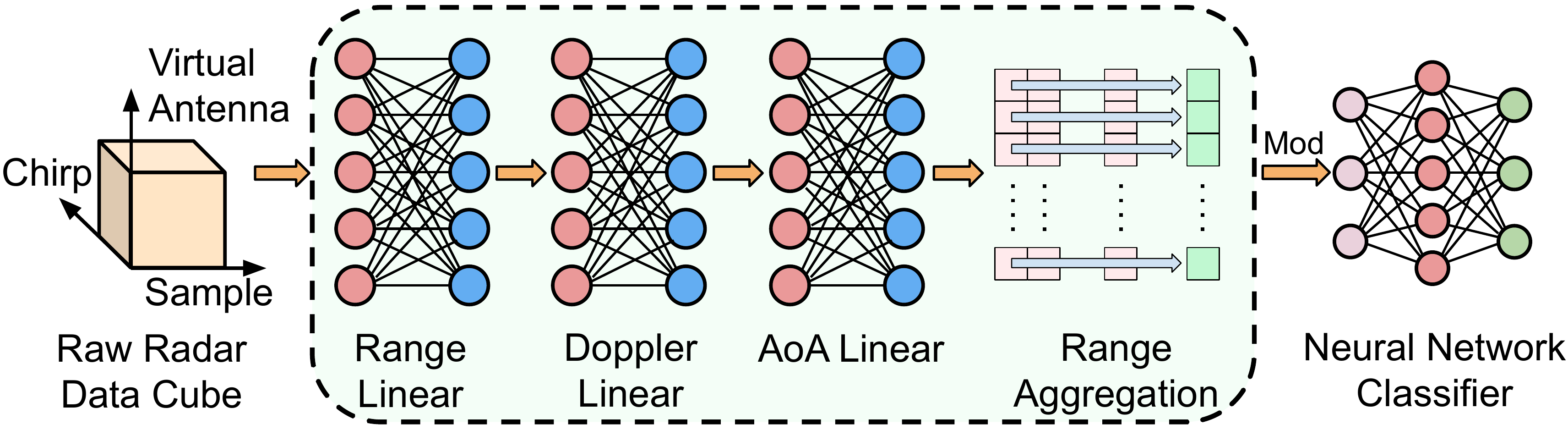}
    \caption{CubeLearn Architecture.}
    \label{fig:learnable}
    \vspace{-10pt}
\end{figure}

The architecture of our proposed CubeLearn module is shown in Fig.~\ref{fig:learnable}. 
The basic idea behind the design is to follow the conventional DFT processing method, as introduced in Sec.~\ref{sec:background}, but in a learnable way. 
As the raw radar signal is complex-valued, we use complex linear layers to directly consume the raw radar signal as input. Instead of processing all the information in the raw radar data cube in one layer, we use a stacked hierarchy of complex linear layers, each layer replacing a certain step in conventional DFT pre-processing (i.e., Range, Doppler and AoA). This is because if we are to use a single complex layer, there would be too many parameters, making the neural network significantly more complex. For example, if we want to replace R-D pre-processing of size (256 x 128), with a single complex linear layer we would need $256*128*256*128 = 1073741824$ complex-valued weights, which is practically unacceptable. However, with two stacked complex linear layers we only need $256*256 + 128*128 = 81920$ complex-valued weights.

As we are using FMCW radar, the first step is to extract Range information from each chirp with a complex linear layer. This can be followed by other complex linear layers to extract Dopper and AoA information for different ranges. We can also aggregate the Doppler and AoA information at different ranges by calculating the sum along the Range axis. To make the proposed CubeLearn module fairly act as a drop-in for the conventional DFT, we calculate the absolute (modulus) value before feeding the processed data into the downstream neural network classifier.

\subsection{CubeLearn Module Weight Initialization}
As conventional DFT pre-processing has been widely adopted and demonstrated universality across different tasks, we use the DFT pre-processing as prior knowledge to initialize our proposed CubeLearn module, and let the neural network adjust the weights to adapt to a specific task through end-to-end training, with a flavor of transfer learning. A conventional DFT of size N applied on the input of size M can be represented as
\begin{equation}
        \hat{x}[k] = \sum_{m=0}^{M-1}e^{-j\frac{2\pi}{N}km}x[m]
    \label{eq:dft}
\end{equation}
where $x$ is the input vector, $M \leq N$, $k=1,2,...,N$, and $j$ is the imaginary sign.

DFT is a linear transformation from time domain to frequency domain, which can be represented by a complex linear layer with no bias in a neural network. For an input of size $M$ and an output of size $N$ ($N \geq M)$, we have:

\begin{equation}
    output[k] = \sum_{m=0}^{M-1}w(k,m)input[m]
\end{equation}
as a result, we initialize the weights as 
\begin{equation}
    w(a,b) = e^{-j\frac{2\pi}{N}ab}
\end{equation}
then we have
\begin{equation}
    output[k] = \sum_{m=0}^{M-1}e^{-j\frac{2\pi}{N}km}input[m]
\end{equation}
which exactly represents DFT. The bias of the complex linear layers are set to 0.

For Doppler and AoA transformation layers we also need to shift the upper half and the lower half of the  weights, since we want zero speed and zero AoA (i.e. in front of the sensor) initially to be in the middle of the learned heatmap. This is because for continuous movements, for example from negative radial speed to positive radial speed, we want the reflection in the heatmap to be continuous as well, rather than jumping from one end to the other.

Note that though we initialize the complex layers as DFT, through end-to-end training the transformation is very likely to no longer be Fourier Transform once trained.

\subsection{Effect of CubeLearn Module}

To better show the effects of our proposed CubeLearn module compared to the conventional DFT pre-processing and validate our assumptions, in this part we demonstrate how the proposed CubeLearn module adaptively adjust the weights, by visualizing the output of both the conventional DFT pre-processing module and the CubeLearn module through the training process. 
We use t-SNE, a widely adopted method for dimension reduction and data visualization \cite{van2008visualizing}. Without loss of generality, we use D-T pre-processing + 2DCNN classifier model on Hand Gesture Recognition task here as an example. Detailed definition of the task as well as the description of the data is introduced later in Section~\ref{sec:experiment_setup}. There are 12 hand gestures in total, however, to make the figures more clear, we only visualize the pre-processed output of 6 gestures. 

\begin{figure}[htbp]\centering
\subfigure[Conventional DFT]{
\begin{minipage}{0.2\textwidth}\centering
\includegraphics[width=\textwidth]{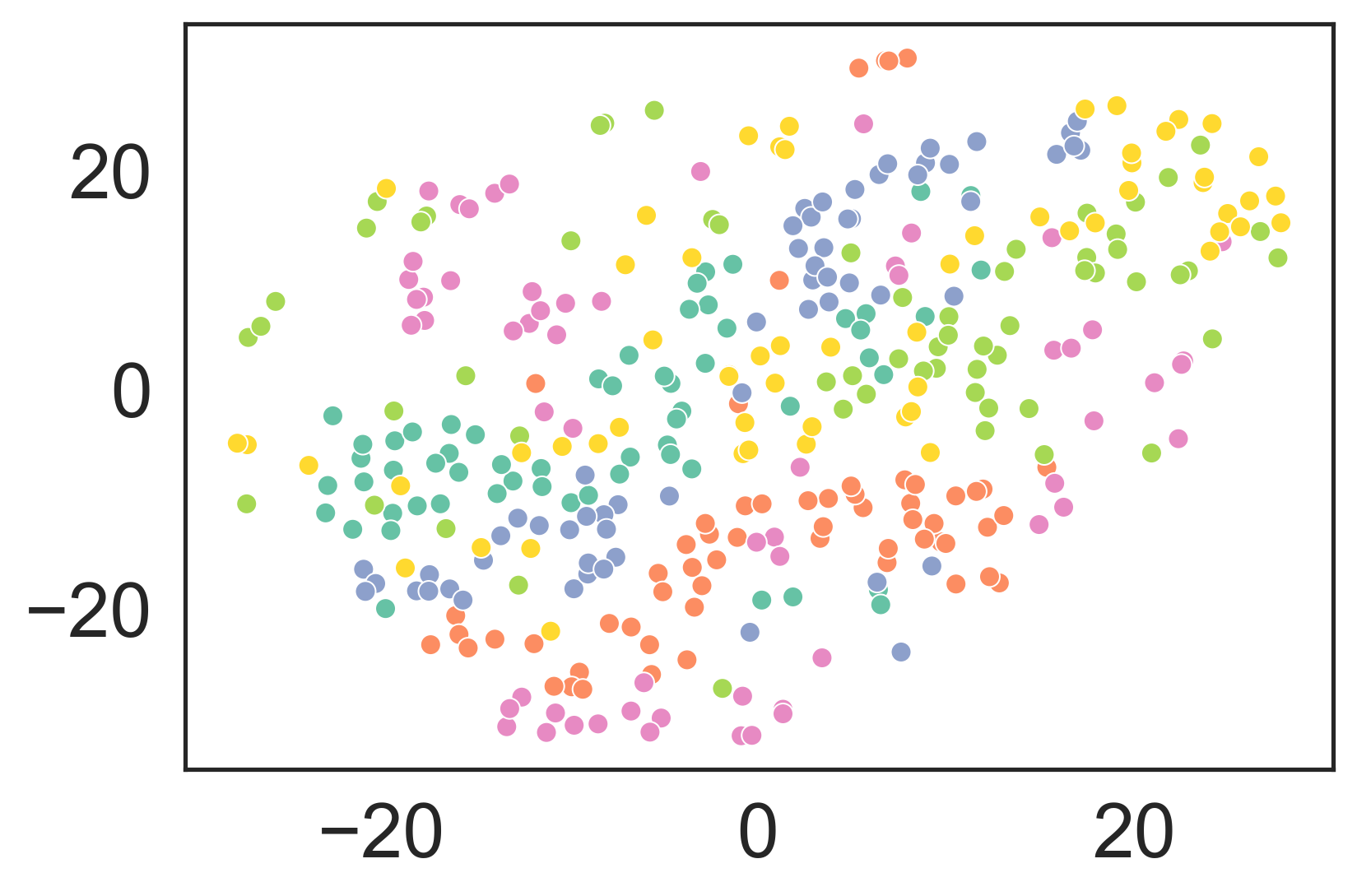}
\end{minipage}
}
\subfigure[CubeLearn, Epoch = 1]{
\begin{minipage}{0.2\textwidth}\centering
\includegraphics[width=\textwidth]{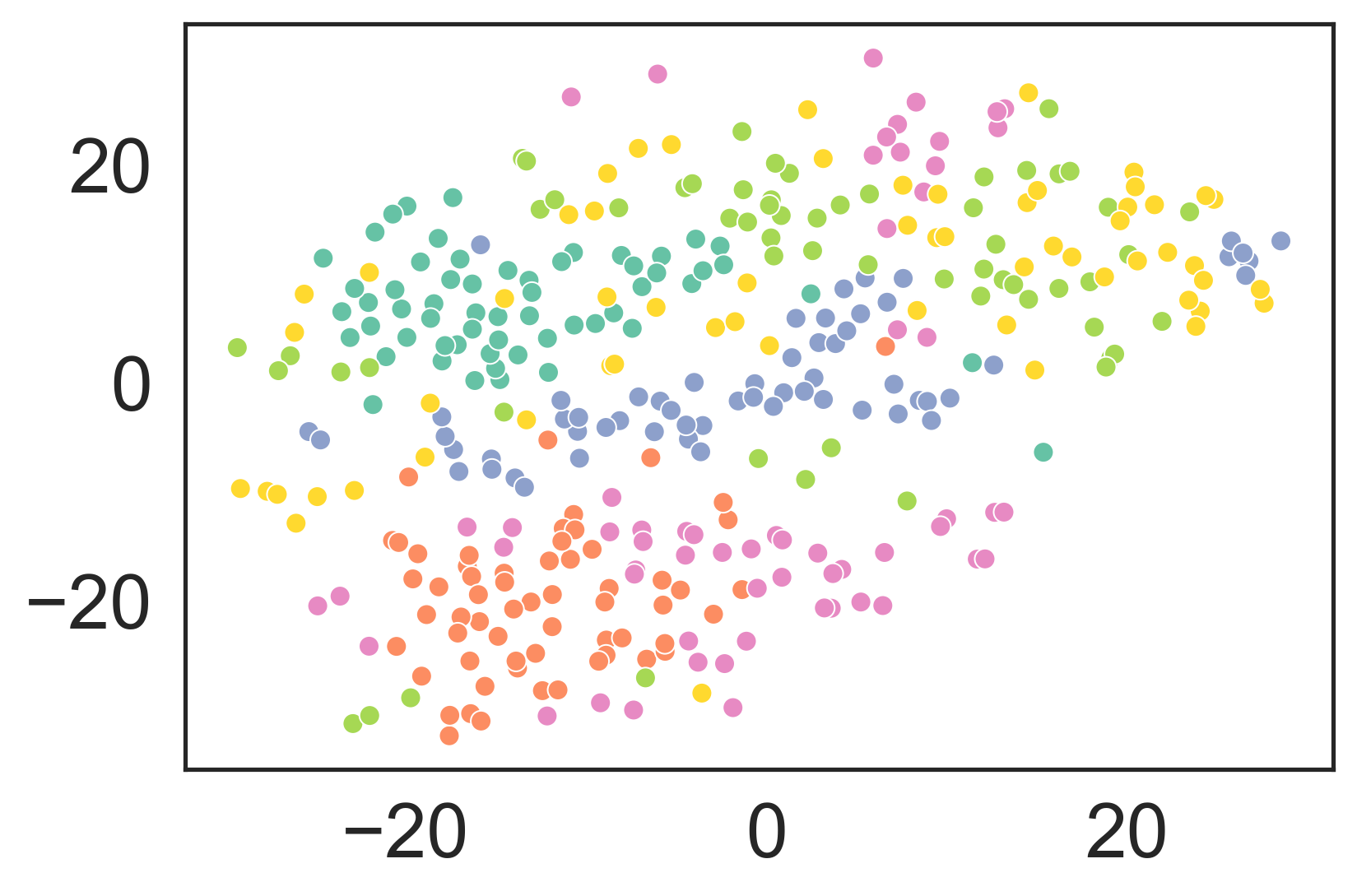}
\end{minipage}
}
\subfigure[CubeLearn, Epoch = 4]{
\vspace{-10pt}
\begin{minipage}{0.2\textwidth}\centering
\includegraphics[width=\textwidth]{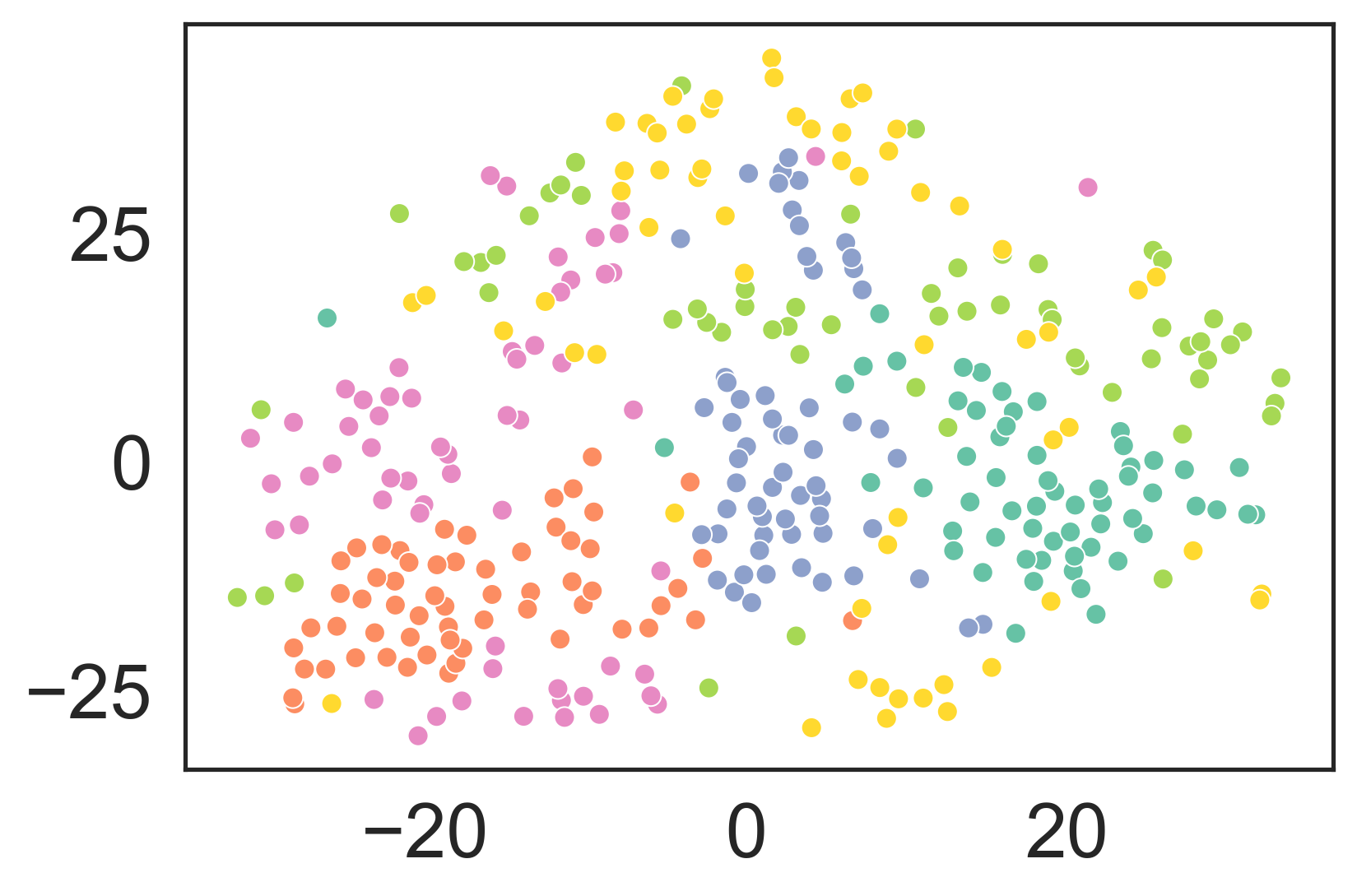}
\end{minipage}
}
\subfigure[CubeLearn, Epoch = 15]{
\vspace{-10pt}
\begin{minipage}{0.2\textwidth}\centering
\includegraphics[width=\textwidth]{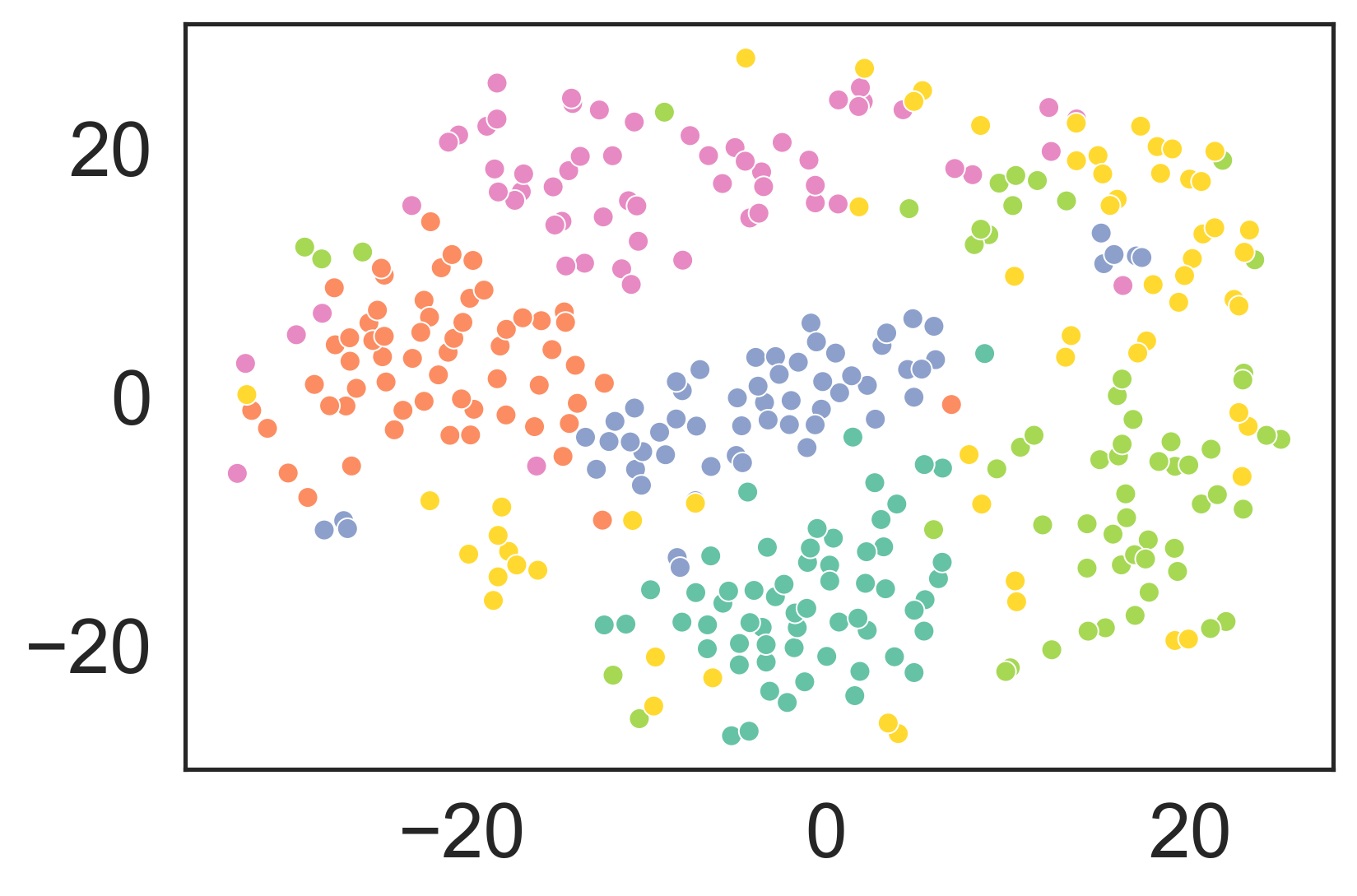}
\end{minipage}
}
\vspace{-10pt}
\caption{t-SNE visualization of the extracted features.}
\label{fig:tsne}
\vspace{-10pt}
\end{figure}

In Fig.~\ref{fig:tsne}, we first visualize the output of the conventional DFT pre-processing in  Fig.~\ref{fig:tsne}~(a), then We visualize the change of the output of the proposed learnable D-T pre-processing module in Fig.~\ref{fig:tsne}~(b)-(d).
From the Fig.~\ref{fig:tsne}~(a) we can clearly see that, although in local regions, the output from certain classes appears to be denser, indicating that the conventional DFT extracts some common properties of the same class samples, on the whole, the features from different classes still tend to mix together. This shows that the conventional DFT pre-processing is not ideal for effectively extracting features from the raw radar signal. With the output from conventional DFT pre-processing, it could be quite challenging for the downstream classifier to produce accurate predictions. 

On the other hand, we can see from Fig.~\ref{fig:tsne}~(b)-(d), with the use of the CubeLearn module, the extracted features of different classes become more distinguishable as the training proceeds. This shows that during end-to-end training, the optimizer is able to adjust the weights in the CubeLearn module to produce more informative outputs. As the outputs of the pre-processing module become more distinguishable, it greatly reduces the burden of the downstream classifier, and makes the task of achieving a higher accuracy simpler.

However, the strong feature extraction ability of the CubeLearn module may overfit to the training data under some cases and produce worse prediction accuracy on `out-of-set' subjects. In later sections, we provide a further evaluation and discussion of the strengths and limitations of the proposed CubeLearn module. 
\section{\label{sec:experiment_setup}Experiment Setup}
We conduct extensive experiments on three most commonly seen interaction tasks with mmWave radar, including hand gesture recognition, arm gesture recognition and human activity recognition. As there is no existing public dataset that contains raw complex radar data available for the above tasks, we collected our own dataset for evaluation. We try to follow existing publications on experiment settings and gesture/activity set designs. To understand the evaluation results, we provide the design of tasks, hardware settings, and software implementation details in this section. 

\subsection{Experiment Design}
In this part we introduce our design of the hand gesture set, arm gesture set and activity set in our experiment.

\begin{figure}[htbp]\centering
\begin{minipage}{0.3\textwidth}
    \includegraphics[width=\textwidth]{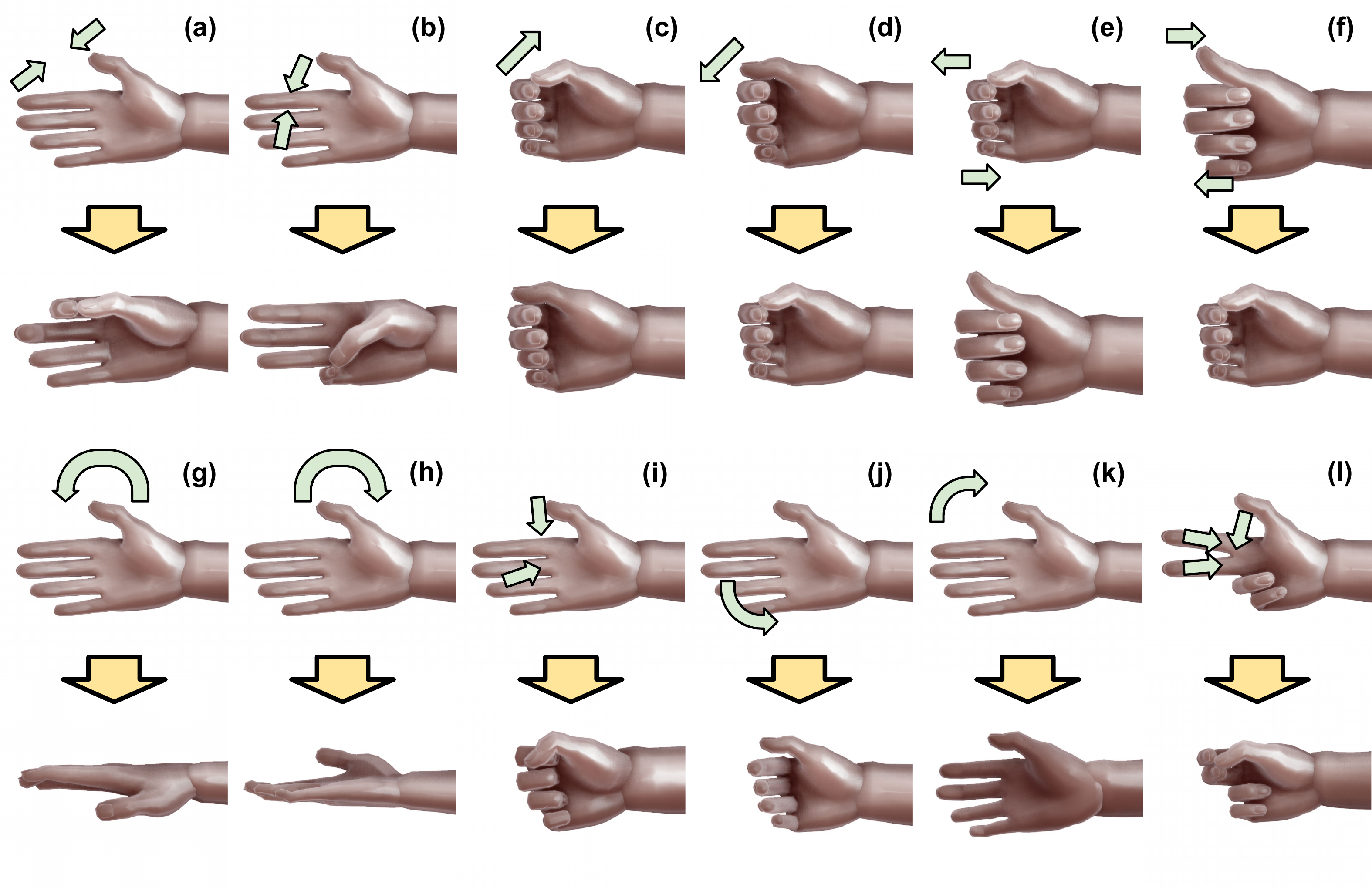}
    \vspace{-20pt}
    \caption{Hand Gesture Set Design.}
    \label{fig:hand_gesture_set}
\end{minipage}
\begin{minipage}{0.15\textwidth}
    \vspace{10pt}
    \includegraphics[width=\textwidth]{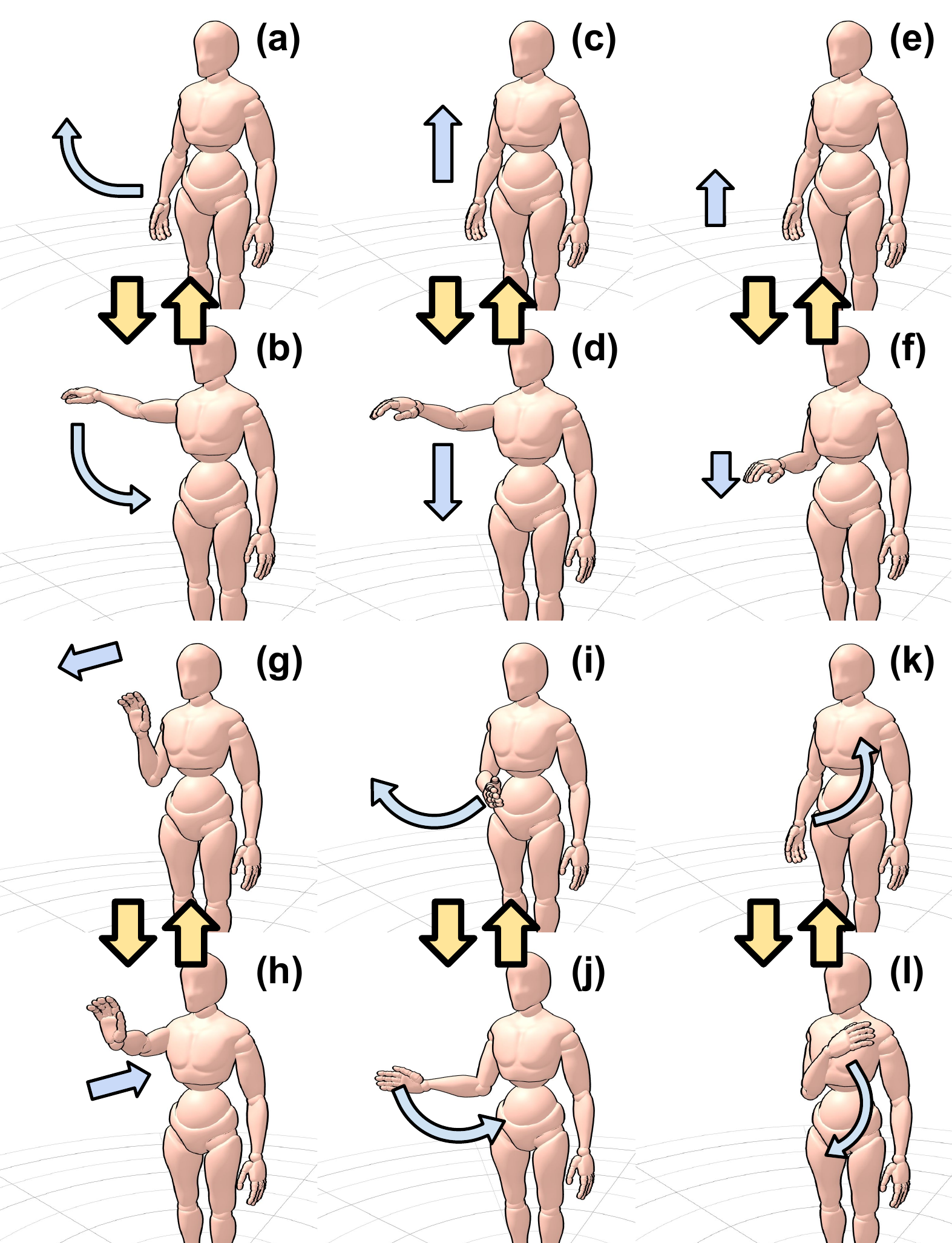}
    \vspace{-20pt}
    \caption{Arm Gesture Set Design.}
    \label{fig:arm_gesture_set}
\end{minipage}
\vspace{-10pt}
\end{figure}

\subsubsection{Task 1: Hand Gesture Recognition (HGR)}

For hand gesture recognition, we follow the gesture set design of several previous works \cite{wang2016interacting, wu2018dynamic, 9381994} and define our gesture set used in this work. Our gesture set design is shown in Fig.~\ref{fig:hand_gesture_set}. The hand gesture set contains: (a) Pinch Index; (b) Pinch Pinky; (c) Slide Right; (d) Slide Left; (e) Rub Forward; (f) Rub Backward; (g) Rotate Left; (h) Rotate Right; (i) Fist; (j) Swipe Left; (k) Swipe Right; (l) Zoom Out. The palm is facing the sensor at the beginning of each gesture, and the gestures are based on the right hand and are performed 20cm away in front of the sensor, as shown in Fig.~\ref{fig:experiment_settings}(b). 

Note that our hand gesture set design is much more difficult than the hand gesture sets in most of the previous works on hand gesture recognition with mmWave radar, as our gesture set only involves hand movement, mostly fingers, and does not include forearm movement.

\subsubsection{Task 2: Arm Gesture Recognition (AGR)}

We design 12 commonly used gestures based on several previous works on radar arm gesture recognition \cite{liu2020real, palipana2021pantomime}, as shown in Fig.~\ref{fig:arm_gesture_set}. The gestures only involve right arm gestures and are performed 1.5m away in front of the radar, as shown in Fig.~\ref{fig:experiment_settings}(c). The arm gesture set contains: (a) Side Lift; (b) Side Down; (c) Front Lift; (d) Front Down; (e) Forearm Lift; (f) Forearm Down; (g) Push; (h) Pull; (i) Swipe Right; (j) Swipe Left; (g) Diagonal Lift; (h) Diagonal Down.

\subsubsection{Task 3: Human Activity Recognition (HAR)}

Another commonly seen application is human activity recognition (HAR). Different from HGR and AGR, HAR usually involves continuous movements, which means the activities do not have a certain `start time' and `end time'. Instead of trying to identify each start time as we do in gesture recognition, we simply cut the data into equal-length samples. We also follow the design in previous works \cite{singh2019radhar, ahuja2021vid2doppler,zhang2018real}, excluding those we believe that are not appropriate for participants to perform (e.g., crawling) or those that are too physically demanding (e.g., jumping) for typical smart-home applications. We include the following 6 activities in our dataset: (a) Marching on the spot; (b) Jogging on the spot; (c) Clapping; (d) Waving right hand; (e) Sweeping the floor with the right hand; (e) Rubbing left arm with right hand. The activities are performed 2m away in front of the radar in our data collection, as shown in Fig.~\ref{fig:experiment_settings}(d).

\subsection{Data Collection Setup}
\begin{figure}[h]
    \centering
    \includegraphics[width=0.48\textwidth]{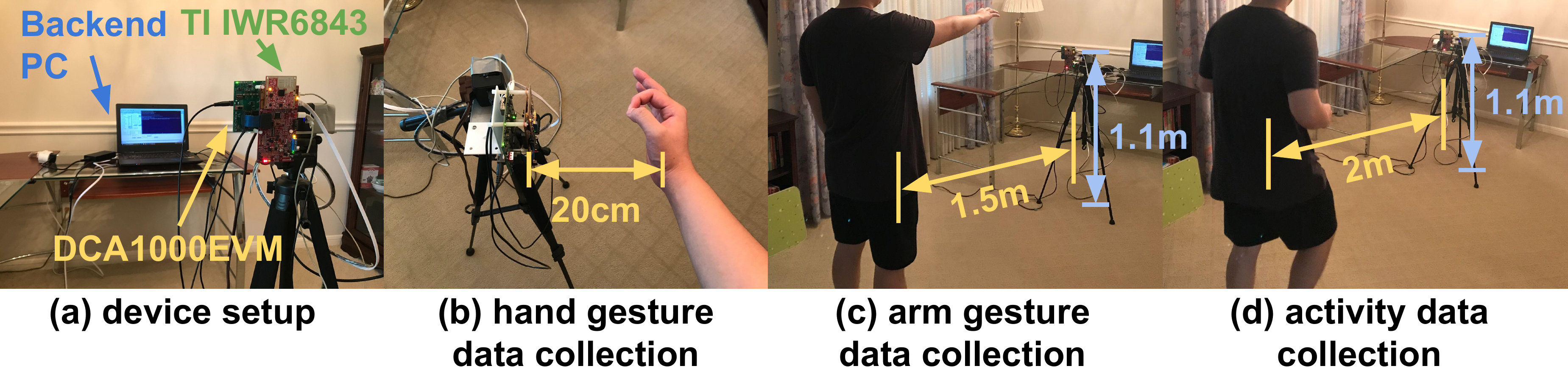}
    \vspace{-10pt}
    \caption{Experiment Settings.}
    \label{fig:experiment_settings}
    \vspace{-5pt}
\end{figure}

In this paper we use a commercial-off-the-shelf Texas Instruments IWR6843ISK mmWave sensor carried by MMWAVEICBOOST, and DCA1000EVM for raw data streaming. The sensor is mounted on a tripod, with a height of around 110cm, as shown in Fig.~\ref{fig:experiment_settings}(a). Raw radar data is streamed and stored in PC continuously. For hand gesture and arm gesture recognition, the starting time of each action is logged and the data corresponding to each action is later extracted from the raw data file according to the starting time with a length of 1 second (10 frames).

We configured the radar with the following parameters: the starting frequency of each chirp is configured to 60.25 GHz with a frequency slope of 60 GHz/ms; the sampling rate is set to $10^7$ samples per second and we take 256 ADC samples at each receiver antenna during a chirp; the period of each frame is 100ms, containing 128 chirp loops where 3 transmitters are activated one by one. 

We invite 8 participants for data collection\footnote{The study has received ethical approval \textit{CS\_C1A\_021\_018}}, including 4 males and 4 females, aged 26 to 55, with height approximately from 160cm to 175cm, and bodyweight approximately from 45kg to 90kg. For each participant, we collected 30 samples per hand/arm gesture, and 1 minute of activity, which is also cut into 30 samples. The background environment of data collection could vary across participants, but the experiment settings are the same. The data from the 8 participants are divided into two parts: 6 people are used as `in-set' users, whose data is used for training and `in-set' testing; the remaining 2 people are `out-of-set' users, whose data is used to test the generalization abilities of the trained model. For `in-set' users, 15 of the 30 samples of each gesture/activity is used for training, 5 used for validation and the remaining 10 used for testing, while for `out-of-set' users, all 30 samples of each gesture/activity are used for testing. 

\subsection{Neural Network Implementation\label{sec:NN_implementation}}
We implement the neural networks with PyTorch. The complex layers are implemented based on an open-source Python module called CplxModule\footnote{\url{https://github.com/ivannz/cplxmodule}}. For Range and Doppler transformations, we configure the output size of the linear layer to be the same as the input size. We only use azimuth AoA information in this work, and for extracting AoA information, we configured the output size to 64 instead of the input size 8, so that the pre-processed data can be better fed into the convolutional layer for further feature extraction. 
For D-A-T and R-D-A-T pre-processing, we use the first half of the raw data on sample and chirp axes because of GPU memory limitation. This does not lose much information on Range and Doppler dimensions, as we will discuss in Section~\ref{sec:raw_size}.

For the implementation of the neural network classifiers, we use the same settings for each model. 
The kernel size for the convolutional layers are set to 3 and the channels are set to 4, 8, and 16, respectively. We apply max-pooling operation of size 2 on each dimension after each convolutional layer excluding the frame dimension.
The LSTM is set to have a hidden size of 512. The output size of the three fully-connected layers is set to 512, 128 and 12/6 (depending on the number of classes), respectively. 

We use the cross-entropy loss to train the network. The neural networks are trained with Adam optimizer. Each model is trained for 30 epochs, and the best weights are saved according to validation accuracy and loss (accuracy is prioritized than loss).

\section{\label{sec:evaluation}Evaluation}
In this section we first evaluate the accuracy of DFT pre-processing based pipelines and CubeLearn based pipelines on HGR task, then extend to AGR and HAR tasks. Some of the pipelines with conventional DFT pre-processing are widely adopted by previous works on related topics, e.g., D-T pre-processing + CNN classifier \cite{dekker2017gesture, wu2018dynamic, ahuja2021vid2doppler, jiang2021recognition, zhang2018real} and R-D-T pre-processing + CNN-LSTM classifier \cite{chen2019gesture, hazra2019short}, etc. As a result, they also serve as a comparison to previous related works. 

We also evaluate the trained models on the `out-of-set' users whose data is not included in the training. Generalization accuracy tends to be lower, as recognition accuracy can be impacted by the shape and size of the user's hand/body, how the user performs the action, as well as the environmental noise. 
This situation can certainly be mitigated by collecting more training data with various users and environments, or adopt transfer learning and life-long learning techniques. As our main aim here is to evaluate the generalization ability of our proposed CubeLearn based methods against conventional DFT pre-processing based pipelines, we do not focus on improving the generalization ability of a specific model/task here.

Besides validating the performance of our proposed CubeLearn module, this section also serves as a comparison between pre-processing and classifier combinations on different tasks.

\subsection{Hand Gesture Recognition Evaluation}
\begin{figure}[ht]
    \vspace{-10pt}
    \centering
    \includegraphics[width=0.47\textwidth]{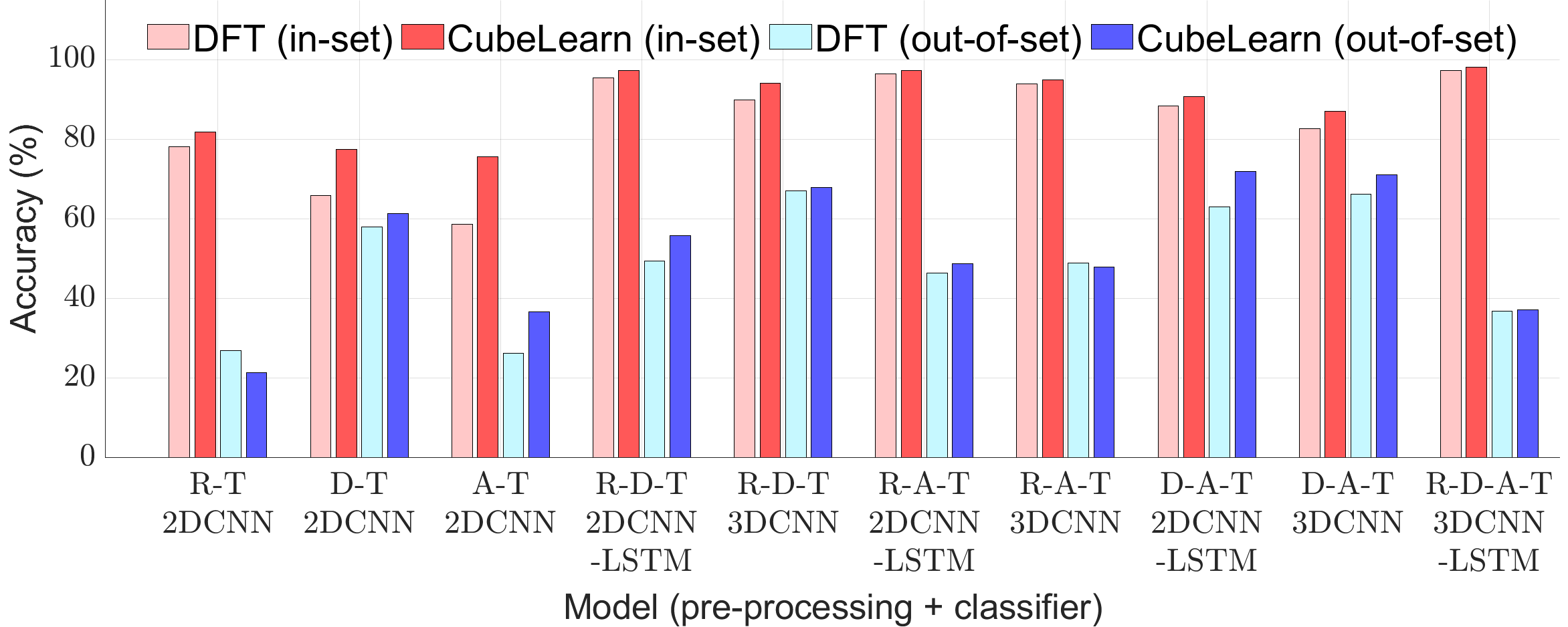}
    \vspace{-10pt}
    \caption{Hand Gesture Recognition Accuracy.}
    \label{fig:hgr_acc}
    \vspace{-10pt}
\end{figure}

We first evaluate our proposed CubeLearn module against conventional DFT on HGR task. The results are shown in Fig.\ref{fig:hgr_acc}.

\subsubsection{Impact of CubeLearn Module}\ 

We can see from the figure that, with CubeLearn, the model is able to have consistently better or at least comparable performance than the DFT pre-processing counterparts. This is especially pronounced on the weaker performing DFT models, e.g., R-T pre-processing + 2DCNN classifier, A-T pre-processing + 2DCNN classifier and the widely adopted D-T pre-processing + 2DCNN classifier, with over 10\% of accuracy improvement for the latter two models. Even for some of the models that already have a reasonably high recognition accuracy ($\geq 90\%$) with conventional DFT pre-processing, our proposed CubeLearn module is usually able to further improve attained accuracies by around $1 \sim 5\%$. These results show that our proposed CubeLearn module is a very powerful tool to boost recognition accuracy on motion recognition tasks with mmWave radar. 

\subsubsection{Impact of the Combination of Pre-processing and Classifier} 
For `in-set' testing, we generally have better performance with higher dimensional pre-processing. R-D-A-T pre-processing based model achieves the best testing accuracy of 97.36\% with conventional DFT and 98.06\% with CubeLearn. And also, two-dimensional pre-processing (R-D-T, R-A-T and D-A-T) have significantly better performance than one-dimensional pre-processing (R-T, D-T, A-T). Higher dimensional pre-processing inherently has more informative features, and the additional information can be utilized to better distinguish between different hand gestures. Also for R-D-T, R-A-T and D-A-T pre-processing, we find that 2DCNN-LSTM classifier behaves better than 3DCNN classifier, probably because the LSTM can better track the temporal relationships between frames in a gesture sample through the explicit latent feature space. 

\subsubsection{Generalization to Out-of-set Users}\ 

When encountering out-of-set users, our proposed CubeLearn can still achieve a better performance than DFT pre-processing for the majority of the models.

Having a closer investigation we can observe that the CubeLearn tends to further increase the generalization ability on more generalizable models, as these models capture more features that represents the underlying physical model, rather than some special properties of the training data.  
For example, in D-A-T + 2DCNN-LSTM classifier, with DFT pre-processing module the model is able to have around 63\% of testing accuracy on `out-of-set' subjects, and the proposed CubeLearn module can improve the accuracy to around 72\%.
    
Comparing across different models, we find that the models with Doppler information have better generalization abilities. This is because Doppler information is invariant to the location where the gesture is performed, as long as the gesture or activity is oriented towards the radar. As CubeLearn can extract more distinguishable Doppler features through a learnable complex linear layer, the generalization abilities of Doppler-based models tend to further improve compared to conventional DFT pre-processing counterparts. 

However, with the strong feature extracting ability of the CubeLearn module, some specific characteristics of the training samples could also be extracted and carried over, which sometimes makes it harder for the model to generalize to previous unseen subjects. For example for R-T pre-processing + 2DCNN classifier, although CubeLearn can attain better test performance with `in-set' users, it generalizes less well than conventional DFT pre-processing, as the Range feature itself is inherently not able to generalize well on `out-of-set' subjects, with the conventional DFT based pre-processing model achieving less than 30\% accuracy. Luckily when solving real-world problems, we would most likely pick models that use features with strong generalization ability, and in these cases CubeLearn module could also be beneficial to the testing accuracy on `out-of-set' users. 

With respect to radar data cube slicing, differently from `in-set' testing, higher dimensional pre-processing does not always lead to better generalization performance. For example, the D-A-T + 3DCNN model has better generalization ability than R-D-A-T + 3DCNN-LSTM model, also because that the R-D-A-T features overfit to the training subjects. This could also be a function of the training set size - a dataset of hundreds of users would likely generalize well, even for higher dimensional models.

\subsection{Generalization to AGR and HAR Task}
\begin{figure}[h]
    \centering
    \includegraphics[width=0.47\textwidth]{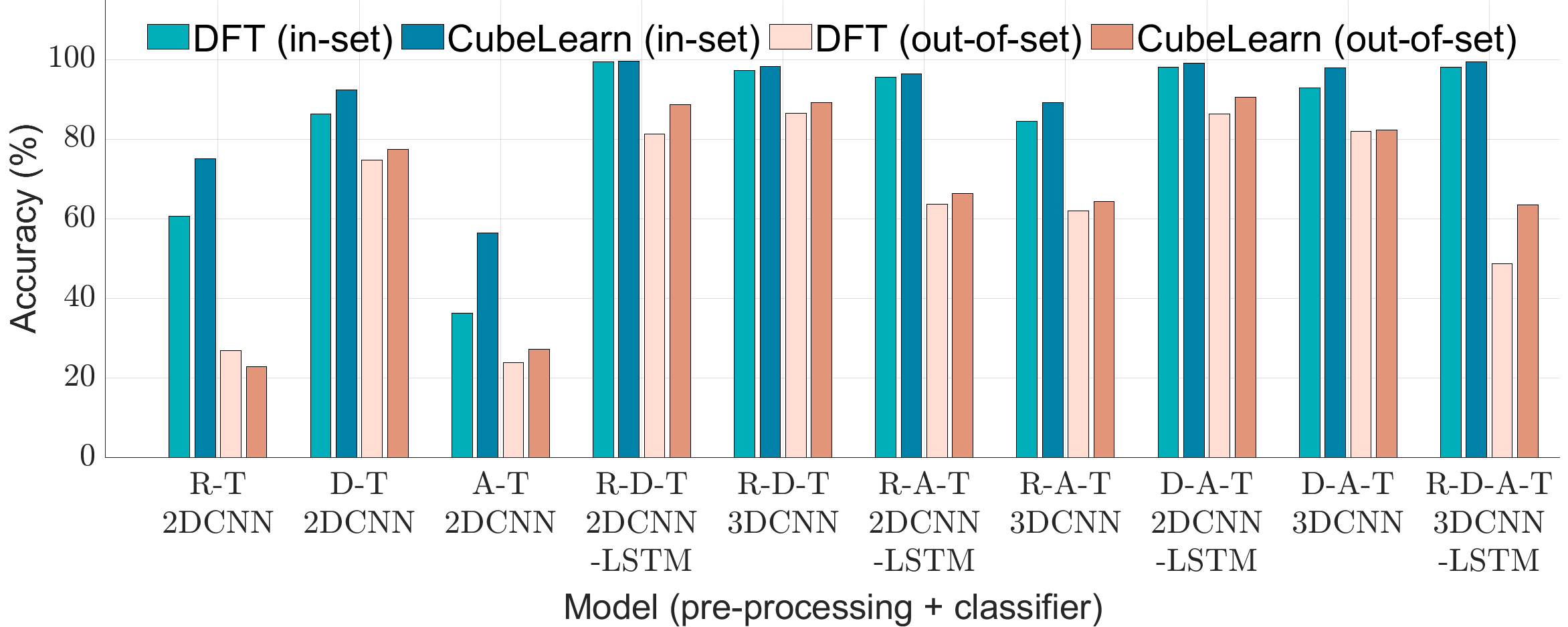}
    \vspace{-10pt}
    \caption{Arm Gesture Recognition Accuracy.}
    \label{fig:AGR_acc}
    \includegraphics[width=0.47\textwidth]{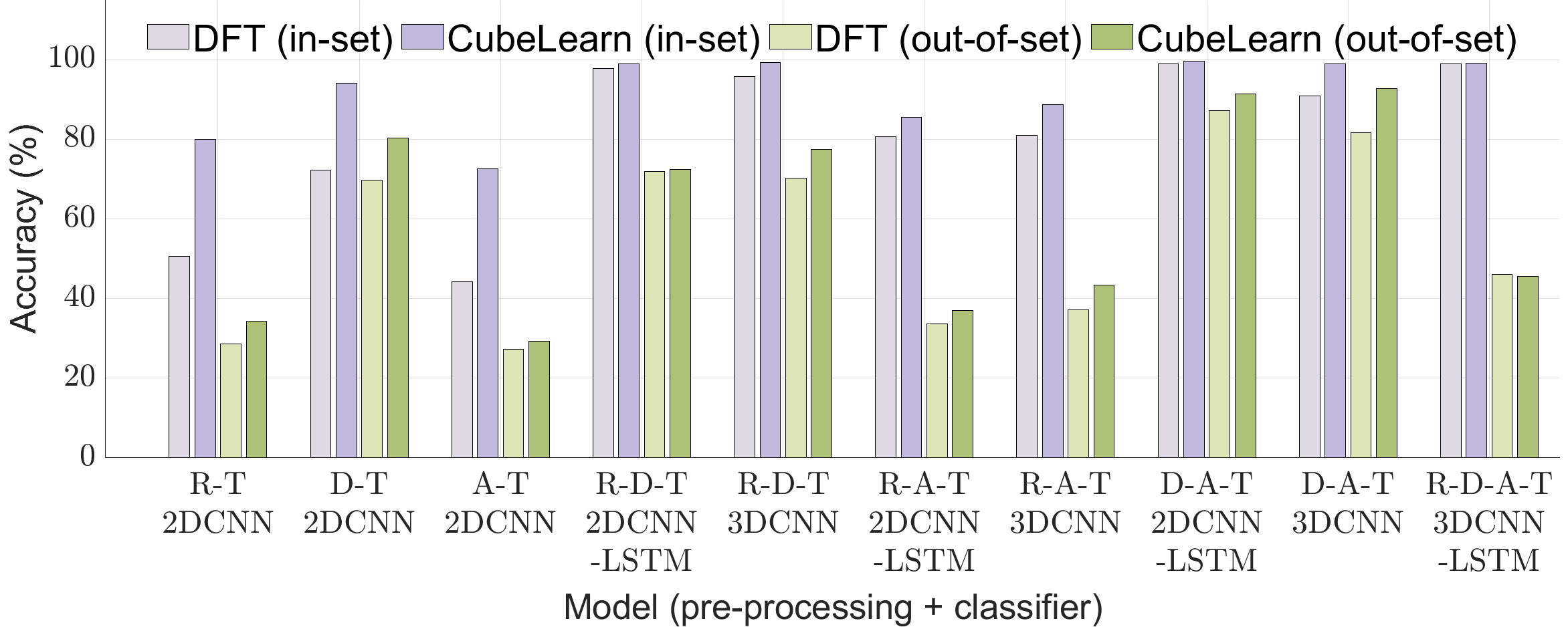}
    \vspace{-10pt}
    \caption{Human Activity Recognition Accuracy.}
    \label{fig:HAR_acc}
    \vspace{-10pt}
\end{figure}

We further evaluate our proposed method against conventional DFT on the other two tasks, and the results are shown in Fig.~\ref{fig:AGR_acc} and Fig.~\ref{fig:HAR_acc}. AGR task is similar to HGR task, where each gesture is represented by a sequence of frames, but it is much easier as the movements are bigger which makes them more obvious to mmWave radar. HAR is fundamentally different as it mainly involves recognizing continuous and repetitive movements. 

From the `in-set' testing result we can see that, the model performances on AGR and HAR are generally better than HGR task. Still, our CubeLearn module can consistently improve the performance of different pipelines, by extracting the most informative features from the raw radar signal cube. The proposed module can be extremely beneficial sometimes, e.g., increasing the accuracies of some of the simpler models (R-T, D-T, A-T) by almost 15\% to 20\%. This is an important finding, as R-T and D-T are the most basic signals which all FMCW radars will support, regardless of the number of antennas. This means that a simpler radar (with respect to hardware configuration) could achieve the performance of a much more sophisticated device, merely through the use of CubeLearn.

Different from HGR/AGR case, in HAR task we find that using LSTM in the downstream classifier does not always yield better performance. For example when using R-A-T pre-processing, the 3DCNN classifier has better performance than the 2DCNN-LSTM classifier. This suggests that the LSTM might be better at recognizing gestures than continuous motion, and for continuous motion recognition, both classifiers are worth trying.

The accuracies for `out-of-set' test are also higher as the tasks themselves are less challenging to generalize due to their more pronounced motion. The accuracies of some models on `out-of-set' subjects are comparable to the accuracies on `in-set' subjects, such D-T + 2DCNN in HAR task. By analyzing the `out-of-set' test results, we can arrive at a similar conclusion as in the HGR task, i.e., our proposed CubeLearn module can be beneficial for most models, and Doppler features are more important for motion recognition than Range or AoA features. 

\subsection{Convergence Analysis}
\begin{figure}[ht]\centering
\vspace{-10pt}
\subfigure[HGR: D-T pre-processing + 2DCNN classifier]{
\begin{minipage}{0.4\textwidth}\centering
\includegraphics[width=\textwidth]{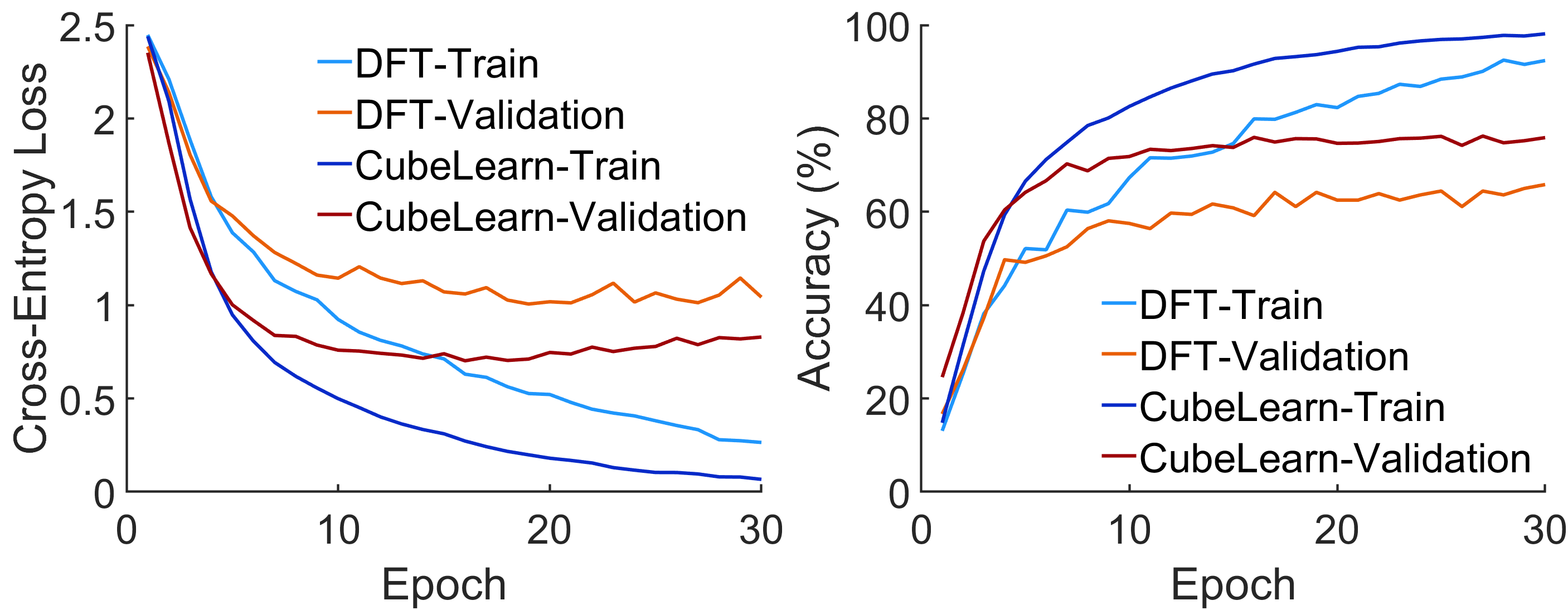}
\end{minipage}
}
\subfigure[HAR: R-D-T pre-processing + 3DCNN classifier]{
\begin{minipage}{0.4\textwidth}\centering
\vspace{-10pt}
\includegraphics[width=\textwidth]{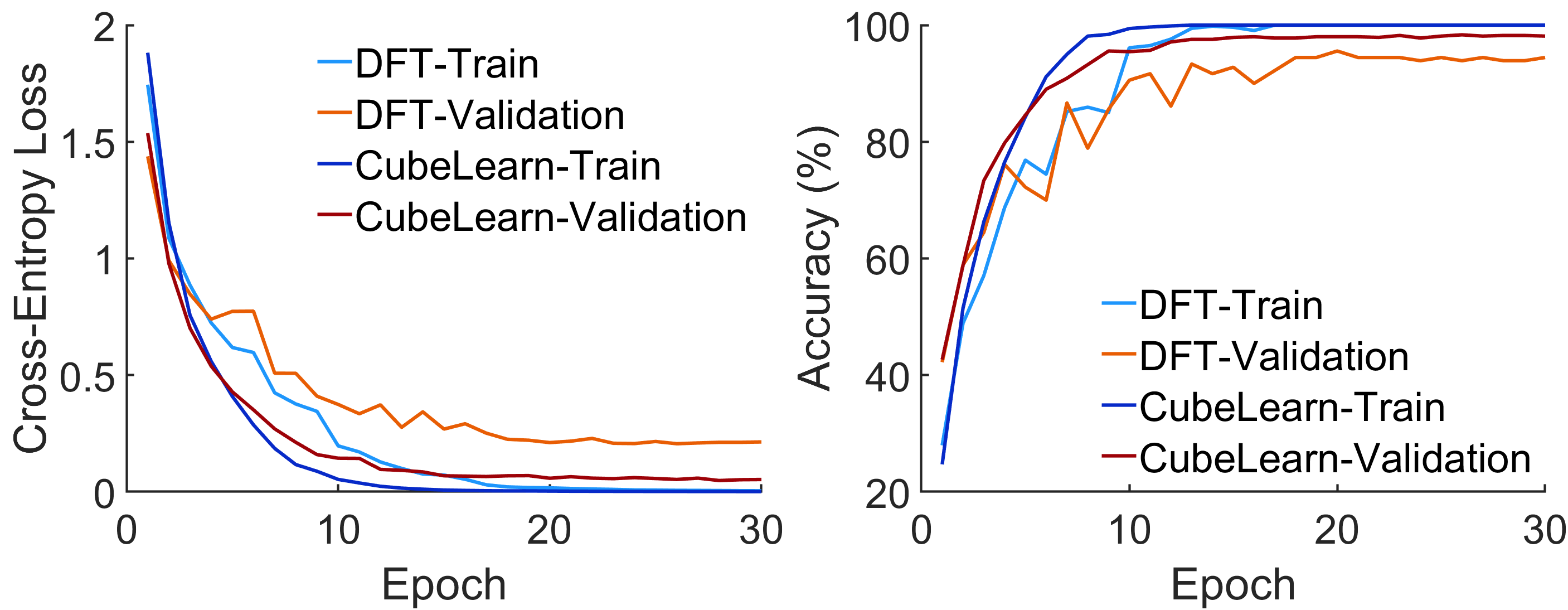}
\end{minipage}
}
\vspace{-10pt}
\caption{Visualization of the training processes.}
\label{fig:training}
\vspace{-5pt}
\end{figure}

We further compare the convergence speed of methods based on DFT pre-processing and CubeLearn. We visualize the training process, including the training accuracy/loss and the validation accuracy/loss, for D-T pre-processing + 2DCNN classifier on HGR task and R-D-T pre-processing + 3DCNN classifier on HAR task in Fig.~\ref{fig:training}. In both cases, the model with CubeLearn is able to converge in fewer epochs, typically within 15 epochs, with loss decreasing and accuracy improving much faster. Furthermore, from  Fig.~\ref{fig:training}~(b) we can see that, though the models converge to approximately the same training loss and accuracy, completely fitting the training set, there is a constant validation loss/accuracy difference between the models, which shows the network with the CubeLearn module can actually better learn the physical model behind the task. Note that this shows that the superior performance of the CubeLearn module is not merely because of more trainable parameters, as in this case CubeLearn based model and conventional DFT based model both completely fit the training data, while the CubeLearn based model still yields better performance on the validation/testing set. 
\section{Ablation Study\label{sec:ablation}}
\subsection{Impact of Complex Layer Learning Rate}

\begin{figure}[ht]
\begin{minipage}{0.45\textwidth}
    \vspace{-10pt}
    \includegraphics[width=\textwidth]{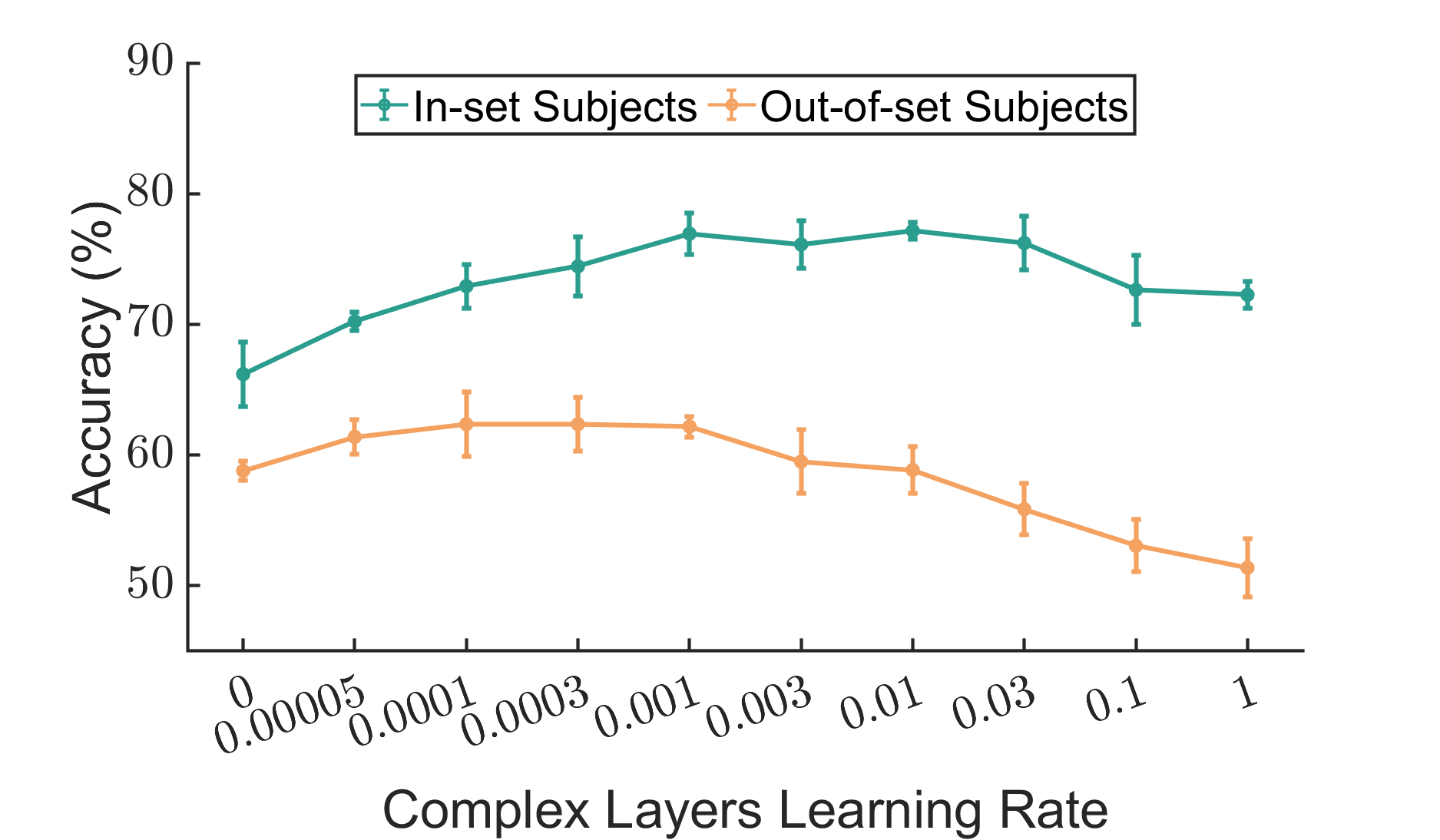}
    \vspace{-20pt}
    \caption{Test accuracy V.S. different learning rates.}
    \label{fig:learning_rate}
\end{minipage}
\vspace{5pt}
\begin{minipage}{0.47\textwidth}
    \includegraphics[width=\textwidth]{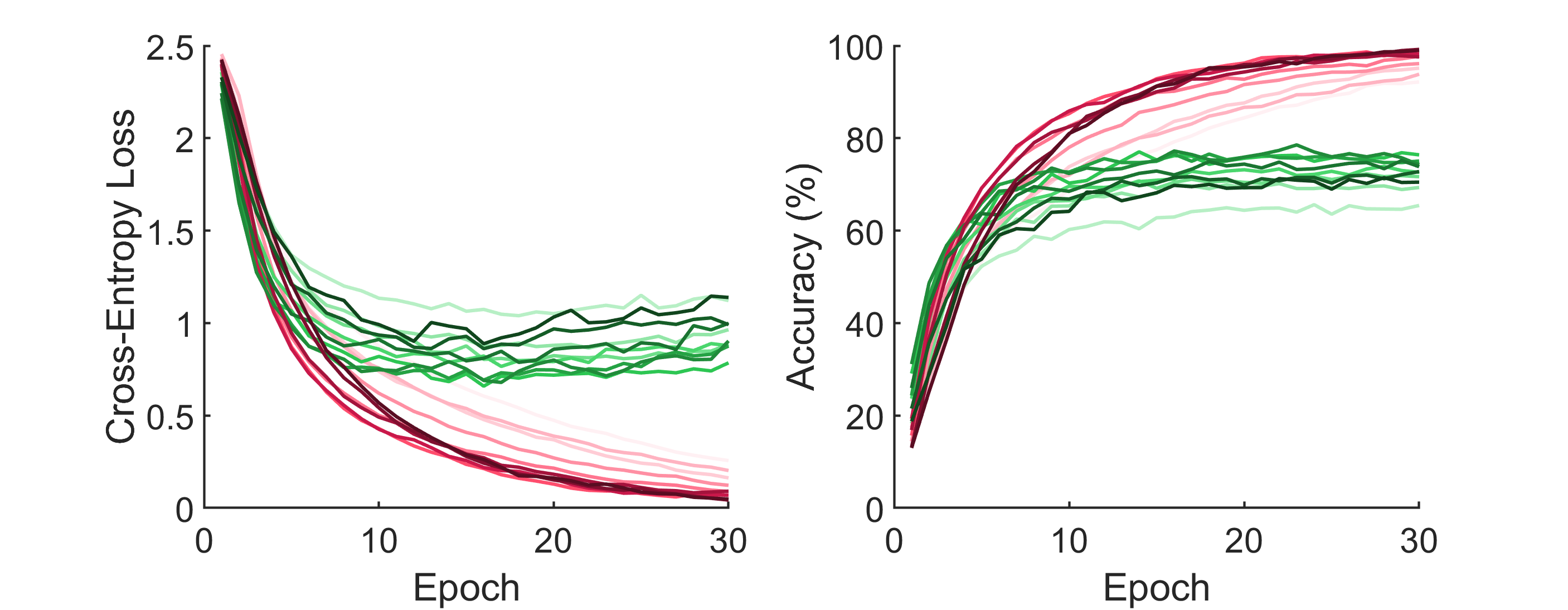}

    \vspace{-20pt}
    \caption{Loss and Accuracy curve. Red: Training; Green: Validation. Darker color represents higher CubeLearn module learning rate. }
    \label{fig:lr_converge}
\end{minipage}
\vspace{-5pt}
\end{figure}

The learning rate of the CubeLearn module controls the speed at which the complex layer weights vary, which can be set differently from the learning rate of the neural network classifier. 
With a higher CubeLearn module learning rate, the weights of the complex linear layers change faster during training - the coupling between the pre-processing and classification networks could cause the system as a whole to converge to a sub-optimal set of weights.
To find the optimal learning rate for the complex layers (or the ratio between the learning rates of the CubeLearn module and the classifier), in this part we study the impact of the CubeLearn module learning rate on the model overall accuracy. We use the D-T pre-processing + 2DCNN classifier model on HGR task here for demonstration. The learning rate of the neural network classifier is set to 0.0003 for all the cases. Each model is run multiple times. The result is shown in Fig.~\ref{fig:learning_rate}. 

From the results we can see that, both in-set and `out-of-set' testing, with the increase of the CubeLearn module learning rate, the model accuracy first increases, then decreases. Note that when the learning rate is 0, the model degenerates to a conventional fixed DFT pre-processing + 2DCNN classifier model. With the learning rate from 0.00005 to 0.0001, the optimizer is able to adjust the weights of the CubeLearn module to achieve better model accuracy, for `in-set' and `out-of-set' subjects test. With the learning rate further increasing, the model is likely to overfit to the training subjects, and even to the training samples. As shown in the figure, the generalization accuracy begins to decrease when the learning rate of the CubeLearn module is larger than 0.001, which shows that the model is overfitted to the training subjects. The `in-set' test accuracy is able to be maintained at a relatively high level until the learning rate reaches 0.03. When the learning rate is larger than 0.03, the model further overfits to the training samples, leading to an accuracy degradation on the `in-set' test as well. 

Fig.~\ref{fig:lr_converge} shows the training and validation loss and accuracy curve of the models with different CubeLearn module learning rates. We can see that for training set, with a higher learning rate the model is able to converge at a smaller training loss, and larger training accuracy. However, a suitable learning rate could make the model converge faster. For validation set, either a learning rate that is too small or too large yields sub-optimal performance, and the smallest loss and highest accuracy can only be achieved with a proper learning rate of the complex layers, which is 0.001 in this case. Note that the optimal learning rate ratio between the CubeLearn module and the neural network classifier can vary for different pre-processing and classifier combinations as well as on different tasks. 

\subsection{Impact of Weight Initialization Method}
The initialization of the weights in the stacked complex linear layers affects the model accuracy as well as convergence. In this part we try different weight initialization strategies, including: (1) Log-DFT initialization, where the weights are initialized with Fourier Transform bases more focusing on the low frequency part (\textbf{LDFT}); (2) Default Random initialization (He Initialization\cite{he2015delving}) for both real and imaginary parts of the complex weights (\textbf{Rand}); (3) Random Non-uniform Discrete Fourier Transform bases initialization, where the initialize weights are based on random Fourier Transform bases (sorted) (\textbf{NUDFT}); (4) Target Focus initialization, where the information (e.g., distance, velocity, AoA) regarding the target is known and the bases of the Fourier initialization are generated close to the target frequency with normal distribution (\textbf{TF}).

\begin{figure}[ht]\centering
\vspace{-10pt}
\subfigure[D-T + 2DCNN classifier]{
\begin{minipage}{0.225\textwidth}\centering
\includegraphics[width=\textwidth]{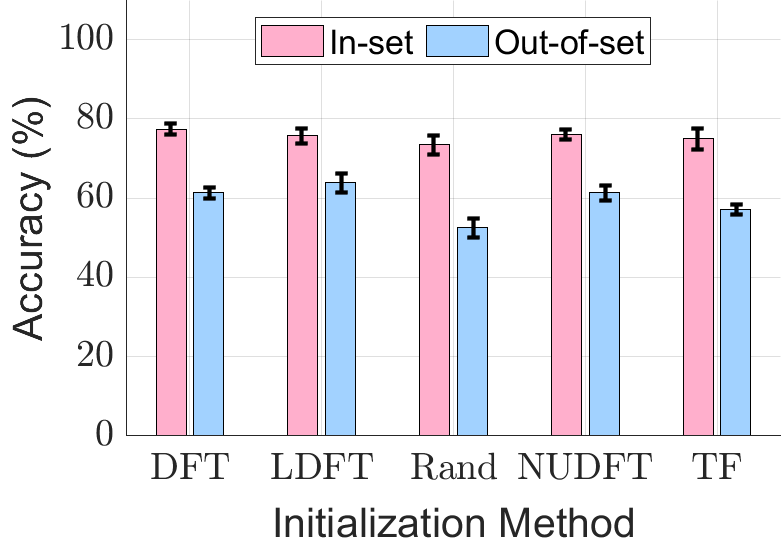}
\end{minipage}
}
\subfigure[R-D-T + 3DCNN classifier]{
\begin{minipage}{0.225\textwidth}\centering
\includegraphics[width=\textwidth]{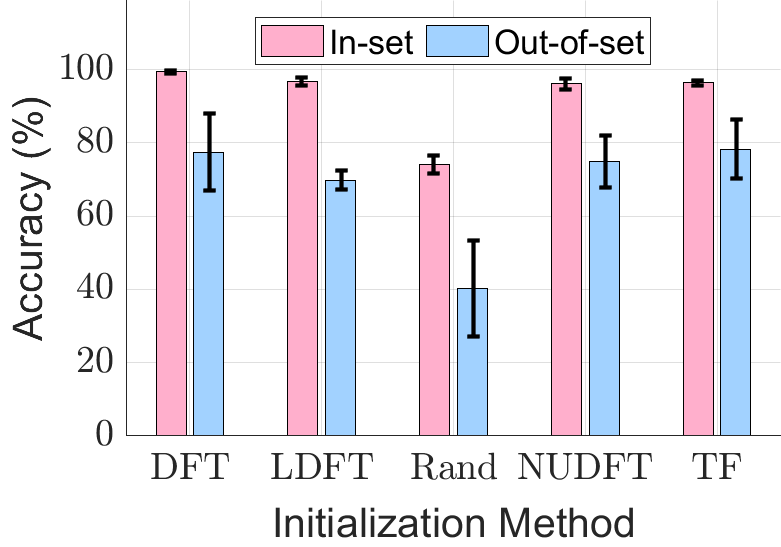}
\end{minipage}
}
\vspace{-10pt}
\caption{Initialization Method Ablation Study.}
\label{fig:initialization}
\vspace{-5pt}
\end{figure}

We compare the complex linear layers initialization methods with two most common models: D-T pre-processing + 2DCNN classifier on HGR task and R-D-T pre-processing + 3DCNN classifier on HAR task. The evaluation result is shown in Fig.~\ref{fig:initialization}, where we can see that the model generally is not very sensitive to the weight initialization method, and is able to converge in each case. 

Log-DFT initialization focuses on the low frequency part, which produces slightly better generalization accuracy for the HGR task, because in the Range domain, the high frequencies correspond to distant targets, which is the background environment in this case. However, we observe a slight performance downgrade for the HAR task, because the user is 2 meters away from the sensor in the HAR task, which is approximately in the middle of the frequency spectrum rather than low-frequency part. With the Log-DFT initialization focusing on the low-frequency part, there are in fact fewer bases initially focus on the target. As a result, Log-DFT might be beneficial for recognizing targets that are close to the sensor, however, this type of initialization needs careful design. 

We find that with random initialization, the model has worse `in-set' and `out-of-set' accuracy. Especially for the HAR case, the accuracy drops to below 80\% and the `out-of-set' generalization accuracy even drops to below 40\%. For HGR the `out-of-set' generalization accuracy also drops significantly. Using random initialization basically means giving up the prior knowledge about radar signal processing, which is not desirable, especially when the number of samples in the training set is limited. 

The Random NDUFT initialization is theoretically quite similar to DFT bases initialization, for the bases are sampled with uniform distribution across the frequency spectrum and re-organized from low frequency to high frequency. However, from the result we see that this kind of initialization is not as stable as uniform DFT bases initialization, with slightly worse average performance.

Target focus initialization is very hard to use in practice, since it requires prior knowledge of the location, velocity and AoA of the target, and a lot of engineering effort. Interestingly, we do not observe better performance even with careful parameter tuning. 

In conclusion, in practice it is reasonable to simply use DFT bases to initialize the weights of the complex linear layers due to their simplicity, high accuracy and stable performance. 

\subsection{Impact of Network Structure}

We further conduct an ablation study of the pre-processing module structure. We also use the D-T pre-processing + 2DCNN classifier model on HGR task and R-D-T pre-processing + 3DCNN classifier model on HAR task here, as they are the most commonly seen models in previous works. Besides, learnable D-T pre-processing can be directly compared to RadarNet \cite{ye2019using} and learnable R-D-T pre-processing can be directly compared to 2D Sinc/Wavelet Filters \cite{stadelmayer2021data}. As mentioned in Section~\ref{sec:related_works}, RadarNet and 2D Sinc/Wavelet Filters are two previously proposed methods for end-to-end radar signal processing. 

We compare the performance of our proposed structure to the following variations: (1) Adding Activation between Complex Linear Layers (\textbf{ModReLU} \cite{arjovsky2016unitary}, \textbf{$\mathbb{C}\text{ReLU}$} \cite{trabelsi2017deep}, \textbf{zReLU} \cite{guberman2016complex}); (2) Learnable Non-uniform Discrete Fourier Transform (\textbf{NUDFT}), where we constrain the learned weights to still represent Fourier Transform bases; (3) Log after Modulus (\textbf{Log}), where the log operation is applied after taking the modulus to represent power spectrum as in \cite{ye2019using, brooks2019complex}; (4) Clutter Removal (\textbf{CR}), where we subtract the chirp average from the raw radar data cube before feeding it into the network; (5) Fully-Complex Network (\textbf{FC}), where we do not take modulus after pre-processing, and use a fully complex version of the downstream classifier; (6) \textbf{RadarNet} \cite{ye2019using}; and (7) 2D \textbf{Sinc} Filters and 2D \textbf{Wavelet} Filters \cite{stadelmayer2021data}.

As RadarNet \cite{ye2019using} is designed for CW radar rather than FMCW radar, we first do Range DFT on the raw radar data, and then apply the pre-processing module introduced in RadarNet for Doppler estimation on each Range bin, then sum along the Range axis. Besides, we use log operation instead of tanh which is suggested in the original paper, as the network does not converge when using tanh in our case. For Sinc and Wavelet Filters \cite{stadelmayer2021data}, we apply it on each frame and use the frame number as an additional dimension, since the radar configuration is different in this work. Besides, we find that the hyper-parameters greatly impact the Sinc/Wavelet Filter performance, so we try different combinations and select the best one. Both RadarNet and Sinc/Wavelet Filters takes real-valued data, and we disregard the imaginary part of the data.

\begin{figure}[h]\centering
\subfigure[HGR: D-T + 2DCNN classifier]{
\begin{minipage}{0.45\textwidth}\centering
\includegraphics[width=\textwidth]{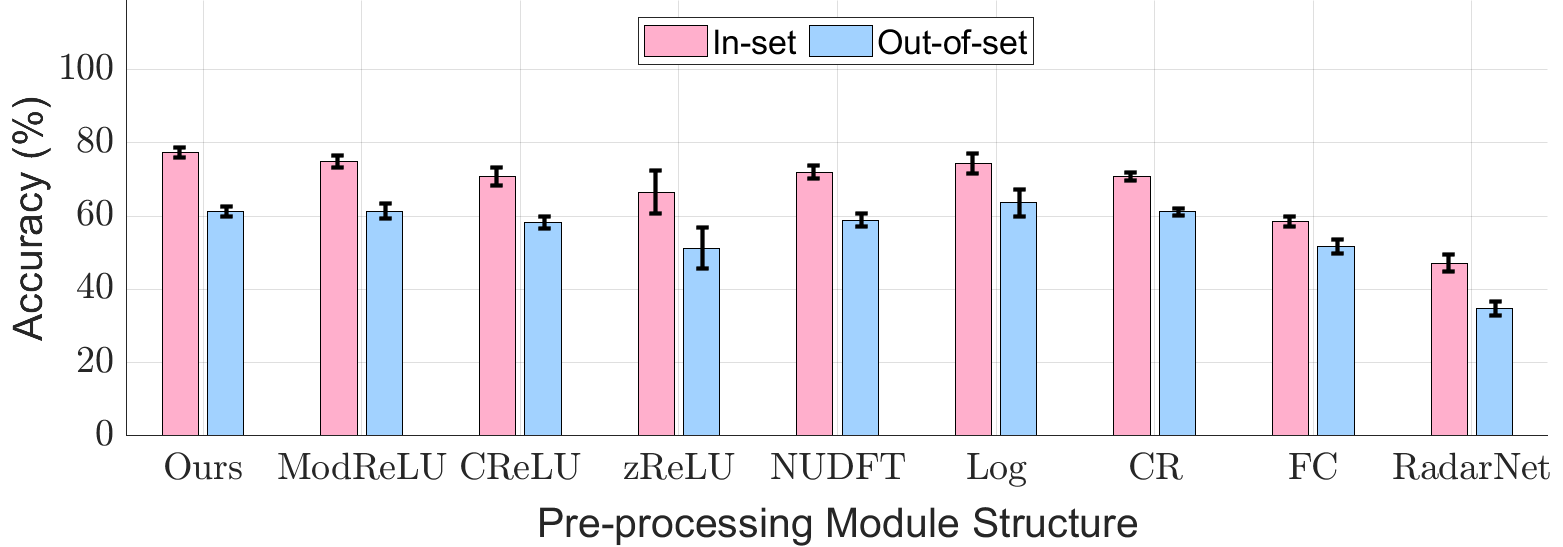}
\end{minipage}
}
\subfigure[HAR: R-D-T + 3DCNN classifier]{
\begin{minipage}{0.45\textwidth}\centering
\vspace{-10pt}
\includegraphics[width=\textwidth]{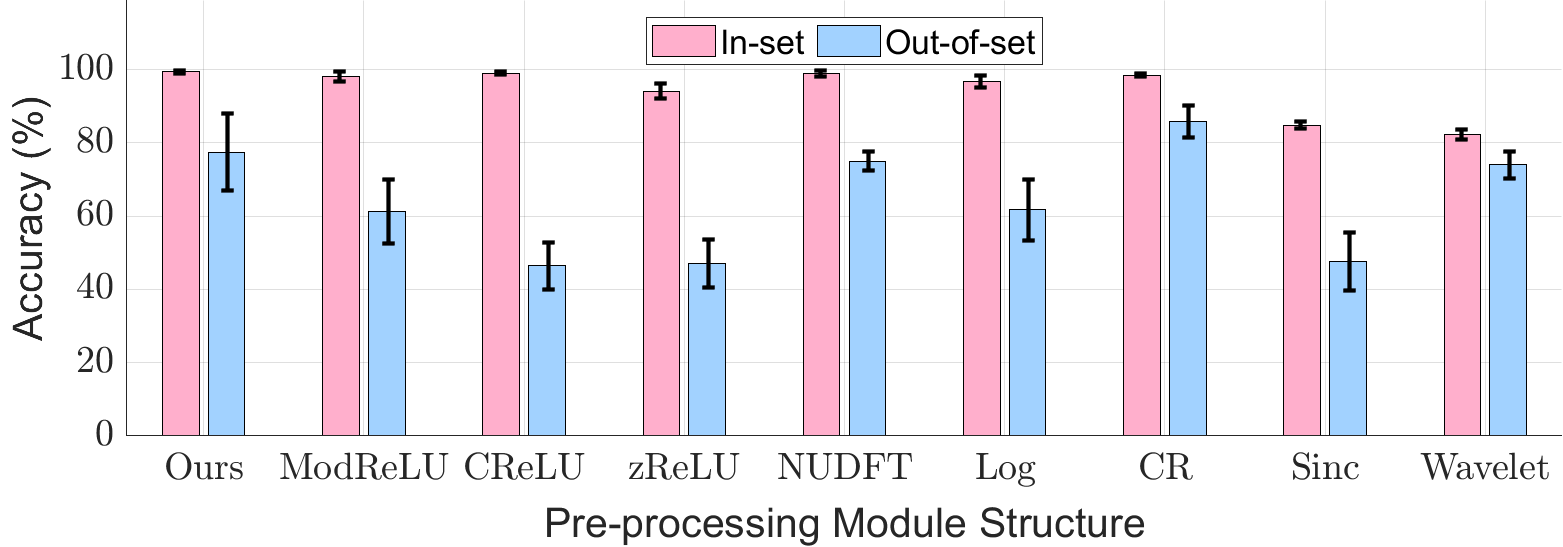}
\end{minipage}
}
\vspace{-10pt}
\caption{Structure Ablation Study.}
\label{fig:structure}
\end{figure}

The result is shown in Fig.~\ref{fig:structure}. We can see that, adding an activation function between the complex linear layers decreases the model performance, especially the generalization ability. This is probably because in the stacked complex linear layers, extracting the Doppler information through the transformation of the latter complex linear layer relies on the output phase information of the previous complex linear layer. Adding a non-linear activation between the complex linear layers, especially $\mathbb{C}\text{ReLU}$ and zReLU, could result in the valuable phase information being altered or lost. 

NUDFT structure produces a slightly worse result than CubeLearn, since the change of the weights is restricted and not as flexible. As for adding the log operation after the CubeLearn module, we do not observe performance improvement. In fact, for the HAR case, adding the log operation downgrades the generalization ability, probably because the difference of peak magnitude and average magnitude is smaller, making the learned features less distinguishable. 

Clutter removal operation removes reflection of static targets (e.g., background), and make the model learn properties of the moving target, which increases the model generalization ability in HAR case. On the other hand, as it actually cancels out all the static information, including the information of the target, for HGR task where the hand is very close to the sensor and occupying a larger field of view, removing the static information slightly decreases the model performance. Besides, if we are to recognize the static target, e.g., in static gesture recognition, clutter removal will certainly not work. As a result, whether adding clutter removal is beneficial or not depends on the task. For tasks to recognize moving targets that are relatively further away, adding clutter removal could increase the model generalization ability by removing the background reflections. While recognizing partial static or static targets, especially in cases where the target occupies a large field of view and the reflection from the environment has limited impact, adding clutter removal could possibly decrease the performance. 

Using the processing module in RadarNet lowers the accuracy. This is because RadarNet only consumes the real input, so it loses half of the information, which leads to a worse accuracy. 2D Sinc Filters and 2D Wavelet Filters also takes real-valued input, so the accuracy is not as good as other structures. 
We find that compared to our proposed CubeLearn with complex linear layers, 2D Sinc Filters and 2D Wavelet Filters are very easy to become overfitted to the training samples, and are very sensitive to the layer parameters. For `out-of-set' test, we observe that 2D Wavelet Filters have competent generalization abilities, much higher than 2D Sinc Filters.

In conclusion, compared to other variations, our proposed stacked complex linear layers achieve good and stable performance. Adding clutter removal could possibly improve the generalization abilities of the model in some tasks, but needs to be used with caution.

\section{Running Time Analysis\label{sec:complexity}}
The Discrete Fourier Transform is usually implemented with Fast Fourier Transform algorithm, which is very computationally efficient ($\mathcal{O}(n\log{}{n})$). Using the stacked complex linear layers introduces more parameters in the network, which will result in larger complexity and longer inference time. In this part we analyze the running time of the proposed method against classical hybrid baselines on PC and Raspberry Pi. The experiment PC we use features an AMD Ryzen 3800X CPU with 64GB memory, and Nvidia RTX2080Ti graphics card, and the Raspberry Pi we use is Raspberry Pi 4 with 8GB main memory. For DFT based pre-processing, we use `torch.fft' in `pytorch' package for possible GPU acceleration when testing on PC, and use `scipy.fftpack.fft' in `scipy' package on Raspberry Pi since `torch.fft' is not supported on ARM architecture. The structures of the neural networks are discussed in Sec.~\ref{sec:NN_implementation}. Note that the running time reported in this section is based on models for HGR/AGR task.

\begin{table}[ht]
    \centering
    \footnotesize
    \begin{tabular}{c|c|cc|cc}
        \hline
        \multirow{2}{*}{Pre-process} & \multirow{2}{*}{Classifier} & \multicolumn{2}{c|}{PC Time (ms)} &  \multicolumn{2}{c}{Raspberry Pi Time (ms)}\\ 
        &&  DFT&CubeLearn& DFT&CubeLearn\\
        \hline
        R-T & 2DCNN  & 0.99 & 1.25 & 15.00 & 35.71 \\
        D-T & 2DCNN  & 1.05 & 1.65 & 77.41 & 484.13\\
        A-T & 2DCNN  & 1.08 & 1.61 & 39.51 & 165.32\\
        \multirow{2}{*}{R-D-T} & 2DCNN + LSTM  & 1.63 & 2.27 & 625.72 & 1018.70 \\
        & 3DCNN  & 1.14 & 1.75 & 419.31 & 745.10\\
        \multirow{2}{*}{R-A-T} & 2DCNN+LSTM  & 1.62 & 2.21 & 308.38 & 528.02\\
        & 3DCNN  & 1.15 & 1.69 & 222.17 & 357.00\\
        \multirow{2}{*}{D-A-T} & 2DCNN+LSTM & 3.18 & 5.29 & 1486.26 & 2456.30\\
        & 3DCNN & 2.92 & 5.02 & 1626.13 & 2513.14\\
        R-D-A-T & 3DCNN+LSTM & 18.73 & 24.38 & -& -\\
        \hline
    \end{tabular}
    \caption{Running time on PC and Raspberry Pi per sample.}
    \label{tab:complexity_time}
    \vspace{-20pt}
\end{table}

The result is summarized in TABLE.~\ref{tab:complexity_time}. 
Not surprisingly, the computation of complex linear layers is more expensive than FFT, as each layer is actually called multiple times during inference. With GPU acceleration on PC, adding the learnable complex layers would add 47.7\% of inference time for one sample on average (SD=15\%). When running with the ARM CPU on Raspberry Pi 4, the running time for each sample increases 152.7\% on average (SD=163.1\%).
The computational cost grows exponentially with the number of dimensions of the raw data cube used. For example, with D-A-T or R-D-A-T pre-processing, the neural network has to perform complex linear transformation on all three dimensions, which leads to a large increase in inference time. 
As a result, if we want to use the proposed CubeLearn, especially on resource-constrained devices, for efficiency reasons we could first consider models that only use a slice from the raw radar data cube, i.e., either R-D or R-A. However, in practice we can have a smaller input size with our proposed end-to-end network, significantly improving the efficiency and still achieve comparable results, as introduced in the next section. 

\section{Impact of Raw Data Size\label{sec:raw_size}}

\begin{figure}[ht]
\subfigure[HGR: D-T + 2DCNN classifier]{
\begin{minipage}{0.225\textwidth}\centering
\includegraphics[width=\textwidth]{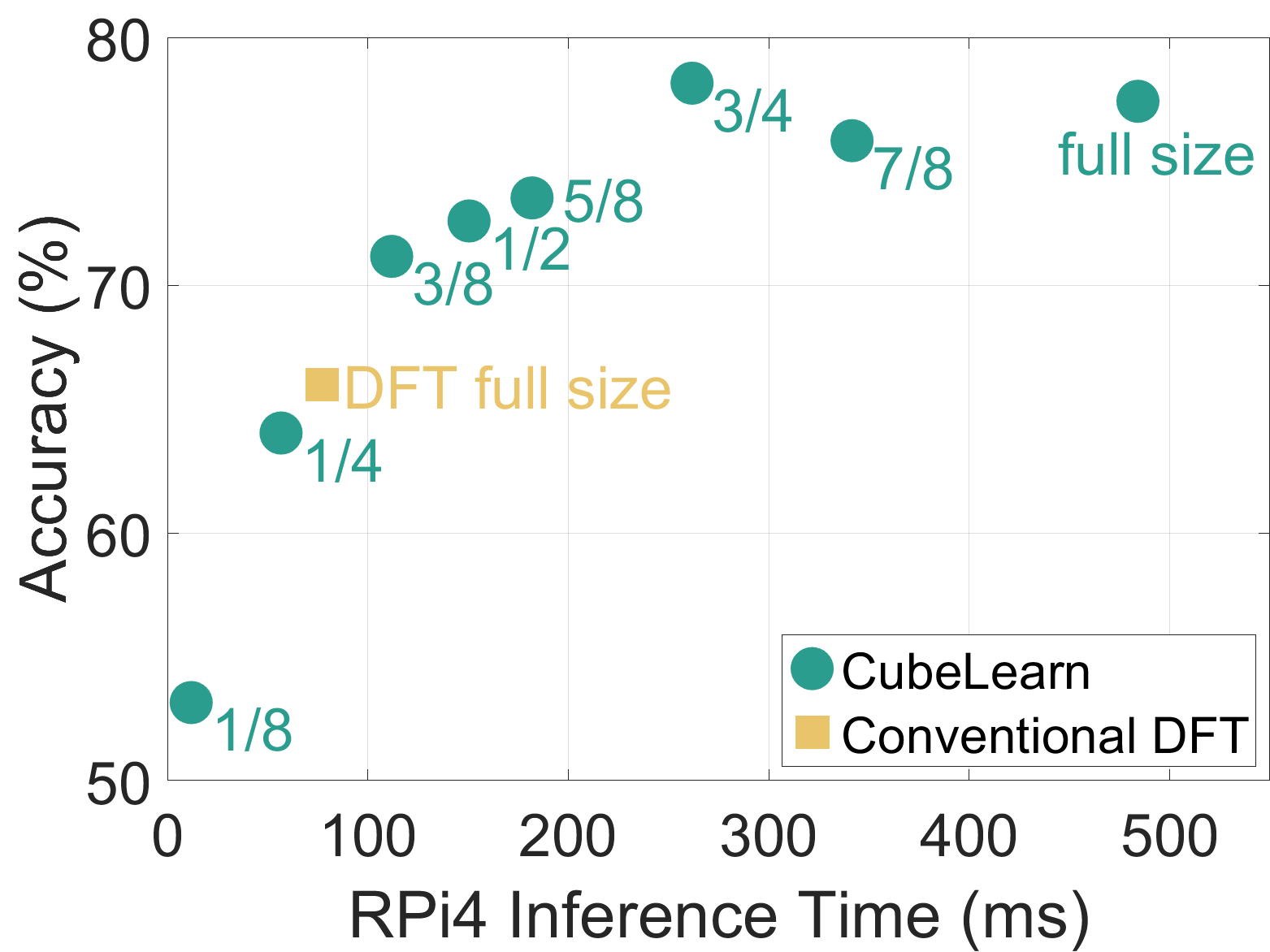}
\end{minipage}
}
\subfigure[HAR: R-D-T + 3DCNN classifier]{
\begin{minipage}{0.225\textwidth}\centering
\includegraphics[width=\textwidth]{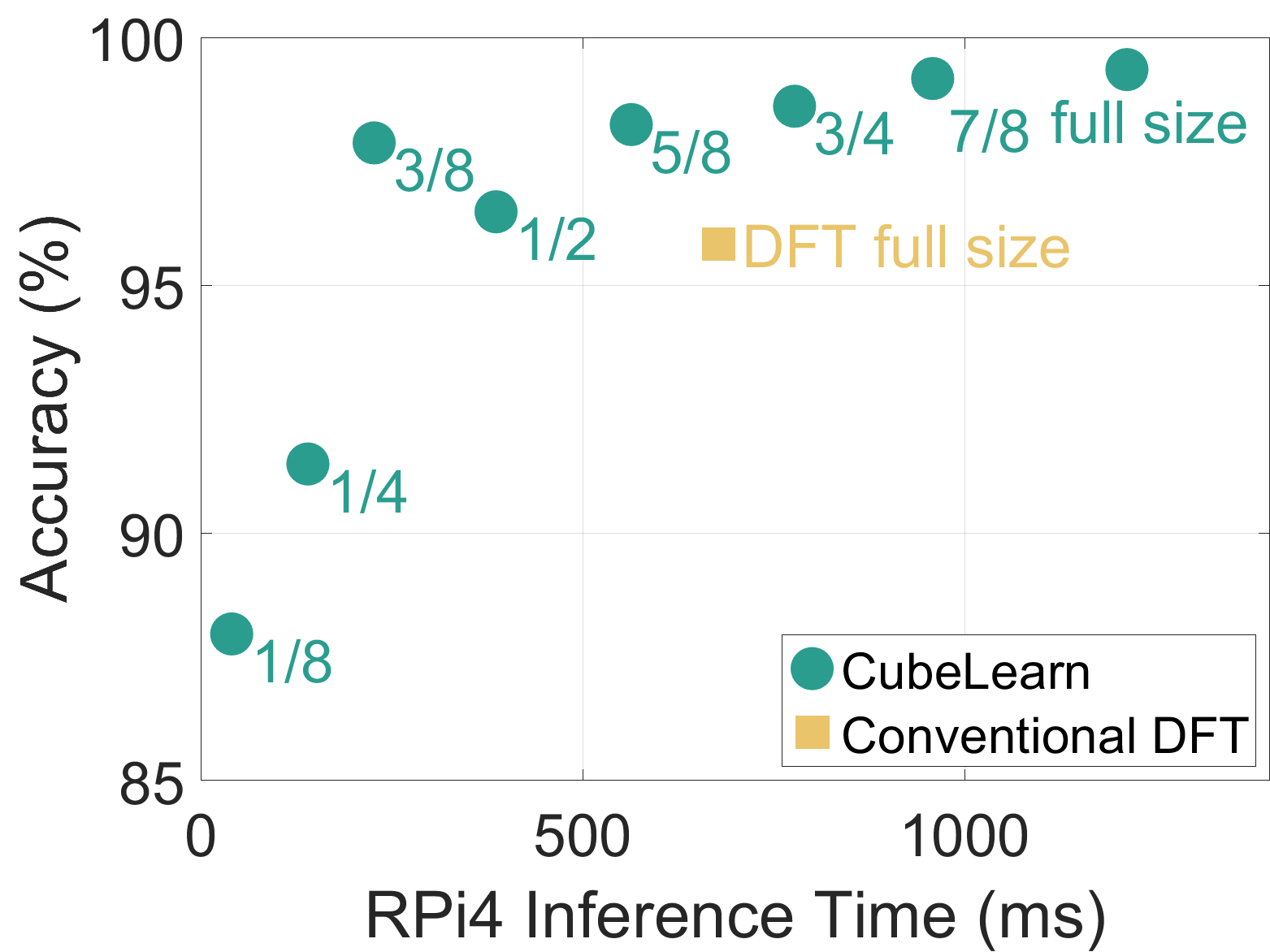}
\end{minipage}
}
\vspace{-10pt}
\caption{Accuracy and inference time with different size of raw data input.}
\label{fig:size}
\end{figure}

The raw radar signal contains repetitive patterns on ADC samples in a chirp, and across chirps in a frame. As a result, lowering the number of samples and chirp number in a frame would possibly still yield a similar level of recognition performance, while making the neural network much more light-weight. In this part we tested the performance of two models with various input sizes: D-T pre-processing + 2DCNN classifier on HGR task and R-D-T pre-processing + 3DCNN classifier on HAR task, as well as the running speed on Raspberry Pi4. Note that the running time of the R-D-T pre-processing + 3DCNN classifier is different from the one reported in Section~\ref{sec:complexity} because each sample lasts 2 seconds for HAR task, instead of 1 second for HGR task which we studied previously.

The input here is actually a series of sample-chirp data slices. For both axes, we try using the first 1/8, 1/4, ..., 7/8 of the data points on each dimension as input. As we have a smaller size of the input, the size of the pre-processing module and the neural network classifier is also smaller, which reduces the computational complexity. The result is shown in Fig.~\ref{fig:size}, the text beside each data point refers the ratio of the samples and chirps we use for each frame. To have a fair comparison, each model is run multiple times. 

We can see that the full-sized input actually contains redundant information. With even 3/8 of the original input size on sample and chirp dimension, we are still able to achieve relatively high performance, better than full sized input with conventional DFT pre-processing. For the HAR case, with 3/8, 1/2 or 5/8 size on each dimension of the original input, the inference time on Raspberry Pi 4 of the CubeLearn based model is smaller than conventional DFT pre-processing based model with full-sized input, but achieves better classification accuracy. Especially with 3/8 of size on both dimensions, the accuracy could reach 98\% while the inference time is approximately 200ms. 

In fact, for specific tasks, we can further reduce the complexity and running time by, e.g., using fewer neurons to represent the transformation output with the initialization of only the Fourier bases corresponding to the distance and velocity of the target. Besides, we can also utilize hardware-level parallel, processing the data of the current frame while receiving the data for the next frame. As a result, we believe that with an optimized implementation as future work, our proposed end-to-end model would execute in real-time on resource constrained devices, with superior performance to the DFT front-end.

\section{\label{sec:conclusion}Conclusion}
In this paper, we propose CubeLearn, a learnable pre-processing module to replace conventional Discrete Fourier Transform (DFT) pre-processing in mmWave FMCW radar gesture/activity recognition pipelines. We demonstrate that through end-to-end training, our proposed CubeLearn module is able to extract more distinguishable features than conventional DFT pre-processing, which makes it easier for the downstream classifier to make correct predictions and improve the overall model accuracy. We evaluate our proposed method on our own collected dataset of hand gestures, arm gestures and activities. Results show that for all the tasks and different pre-processing/classifier combinations, the classification accuracy can be consistently improved with the proposed CubeLearn module, and the training is able to converge in fewer epochs. We envision our proposed method as a small step towards a universal feature extractor for end-to-end deep learning on raw mmWave radar data. Future works would be focusing on exploring more efficient end-to-end structures and improving the model robustness. 



\bibliographystyle{ACM-Reference-Format}
\bibliography{main.bbl}


\end{document}